\pdfoutput=1

\documentclass[11pt]{article}

\usepackage[final]{acl}

\usepackage{times}
\usepackage{latexsym}
\usepackage{amssymb}
\usepackage{tcolorbox}
\usepackage{multirow}
\usepackage{array}
\usepackage{amsmath} 
\usepackage{enumitem}
\usepackage[normalem]{ulem}
\newcolumntype{C}[1]{>{\centering\arraybackslash}p{#1}}
\newcolumntype{P}[1]{>{\arraybackslash}p{#1}}
\usepackage{arydshln}
\usepackage{pifont}
\usepackage{booktabs}
\usepackage{xcolor}
\usepackage{bigstrut} 
\usepackage[linesnumbered,ruled,vlined]{algorithm2e}

\usepackage[T1]{fontenc}

\newcommand{\modify}[1]{#1}

\usepackage[utf8]{inputenc}

\usepackage{microtype}

\usepackage{inconsolata}

\usepackage{graphicx}
\usepackage{makecell}

\definecolor{meta_prompt_color}{HTML}{d1c1df}
\definecolor{anchor_color}{HTML}{b9c6e7}
\definecolor{feedback_color}{HTML}{f3d8c0}
\definecolor{prompt_color}{HTML}{efc7cc}
\newcommand{\green}[1]{\textcolor[RGB]{26,217,22}{#1}}
\newcommand{\gray}[1]{\textcolor[RGB]{128,128,128}{#1}}

\def \shortname{ERM}
\def \taskmodel{task model}
\def \promptoptimizer{prompt optimizer}

\title{
Efficient and Accurate Prompt Optimization: \\
the Benefit of Memory in Exemplar-Guided Reflection
}

\author{
 \textbf{Cilin Yan\textsuperscript{1}\footnotemark[1]\footnotemark[2]},
 \textbf{Jingyun Wang\textsuperscript{1}\footnotemark[1]},
 \textbf{Lin Zhang\textsuperscript{2}\footnotemark[1]},
 \textbf{Ruihui Zhao\textsuperscript{2}},
 \textbf{Xiaopu Wu\textsuperscript{2}},
\\
 \textbf{Kai Xiong\textsuperscript{2}},
 \textbf{Qingsong Liu\textsuperscript{2}},
 \textbf{Guoliang Kang \textsuperscript{1}\footnotemark[3]},
 \textbf{Yangyang Kang\textsuperscript{3,2}\footnotemark[3]}
\\
\\
 \textsuperscript{1}Beihang University,
 \textsuperscript{2}ByteDance,
 \textsuperscript{3}Zhejiang University
\\
   \texttt{\{clyanhh, wangjingyun0730, kgl.prml\}@gmail.com, zhanglin.hb@bytedance.com }\\
   \texttt{\{zhaoruihui, wuxiaopu, xiongkai.kx, liuqingsong, yangyangkang\}@bytedance.com}
}

\begin{document}
\maketitle

\renewcommand{\thefootnote}{\fnsymbol{footnote}}
\footnotetext[1]{Equal contribution. }
\footnotetext[2]{This work was conducted when Cilin Yan was interning at ByteDance.}
\footnotetext[3]{Corresponding authors.}

\begin{abstract}
Automatic prompt engineering aims to enhance the generation quality of large language models (LLMs).
Recent works utilize feedbacks generated from erroneous cases to guide the prompt optimization.
During inference, they may further retrieve several
semantically related exemplars and concatenate them to the optimized prompts to improve the performance.
However, those works only utilize the feedbacks at the current step, ignoring historical and unseleccted feedbacks which are potentially beneficial. 
Moreover, the selection of exemplars only considers the general semantic relationship and may not be optimal in terms of task performance and matching with the optimized prompt. 
In this work, we propose an 
\textbf{E}xemplar-Guided \textbf{R}eflection with \textbf{M}emory mechanism~(\shortname{})
to realize more efficient and accurate prompt optimization.
Specifically, we design an exemplar-guided reflection mechanism where the feedback generation is additionally guided by the generated exemplars. 
We further build two kinds of memory to fully utilize the historical feedback information and support more effective exemplar retrieval.
Empirical evaluations show our method surpasses previous state-of-the-arts with less optimization steps, \emph{i.e.,} improving F1 score by 10.1 on LIAR dataset,
and reducing half of the optimization steps on ProTeGi.

\end{abstract}
\section{Introduction}
\begin{figure*}[ht]
\centering
  \includegraphics[width=0.98\linewidth]{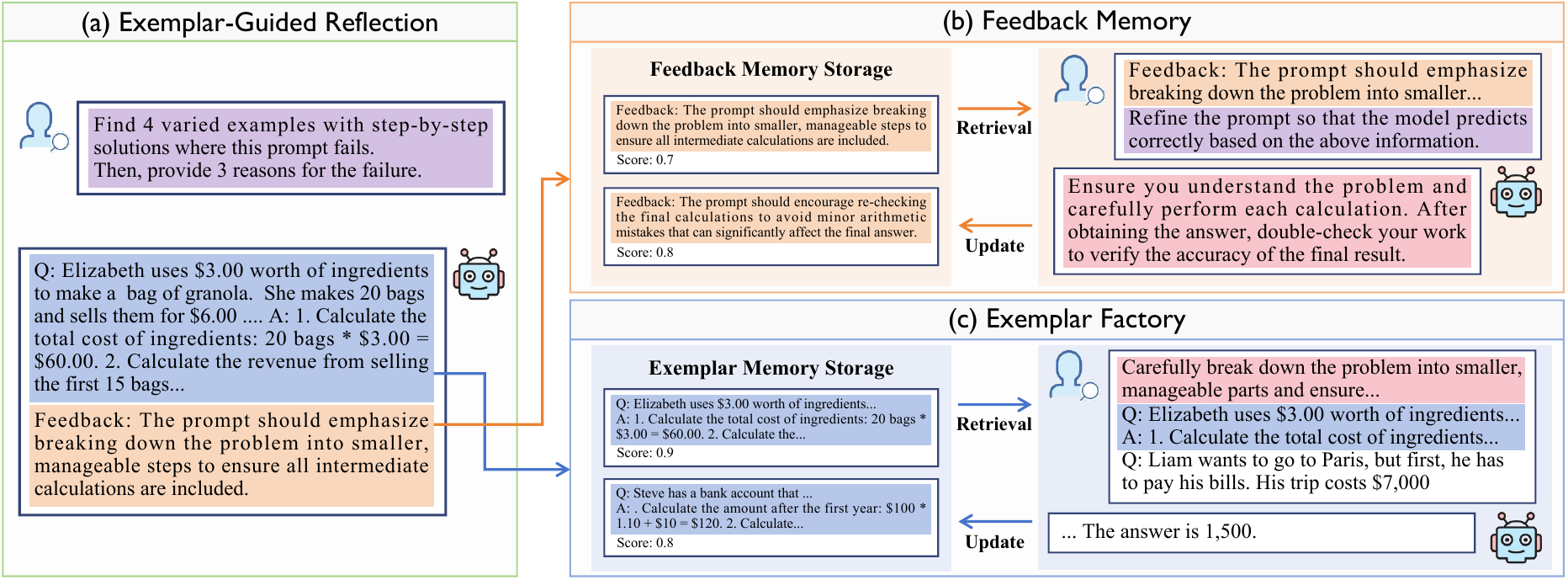}
  \caption{
Feedback-based automatic prompt engineering methods commonly employ a \colorbox{meta_prompt_color}{meta-prompt}, which guides LLMs to evaluate the current case, provide \colorbox{feedback_color}{feedbacks}, and generate refined \colorbox{prompt_color}{prompts}.
In this work, we design an instructive \colorbox{meta_prompt_color}{meta-prompt} to select \colorbox{anchor_color}{exemplars} with detailed solution processes, and generate \colorbox{feedback_color}{feedbacks} for the current case. 
These \colorbox{feedback_color}{feedbacks} are stored in Feedback Memory and periodically retrieved to efficiently guide the optimization of \colorbox{prompt_color}{prompts}.
Additionally, these \colorbox{anchor_color}{exemplars} are stored and assessed in an Exemplar Factory to enhance prediction accuracy.
  }
  \label{fig:demo}
\end{figure*}

Prompt optimization is crucial for enhancing the performance of Large Language Models~(LLMs). 
Even a subtle adjustment to the prompt may lead to an obvious improvement or decline in performance, thereby highlighting the critical role of prompt engineering for LLMs.
Manual prompt engineering demands significant human effort and expert knowledge, while traditional fine-tuning methods~\cite{lester2021power,shin2020autoprompt} heavily rely on substantial computational resources and powerful GPUs.
Therefore, it is necessary to explore automatic prompt engineering, which is 
compatible with black-box APIs (\emph{e.g.,} GPT-4) and does not require extensive resources.

Recently, feedback-based methods~\cite{ye2023prompt, juneja2024task} exhibit promising performance for automatic prompt engineering, which generally  leverage feedbacks generated from failure cases to facilitate the prompt optimization process.
Previous feedback-based methods~\citep{pryzant2023automatic,ye2023prompt} have two main drawbacks. 
Firstly, they throw unselected and historical feedbacks which may benefit the prompt optimization, resulting in more optimization steps to achieve satisfactory performance.
Secondly, during inference, previous methods~\citep{hu2023evoke,juneja2024task} may retrieve several \modify{semantically related exemplars} and concatenate them to the optimized prompt to improve the performance. However, the retrieved exemplars are not optimal without evaluating their influence on the task performance.
Those drawbacks largely constrain both the \modify{efficiency and accuracy} of  
the prompt optimization process.

In this work, we introduce an \textbf{E}xemplar-Guided \textbf{R}eflection with \textbf{M}emory mechanism~(\shortname{}) to achieve efficient and accurate prompt optimization.
Firstly, we propose an exemplar-guided reflection mechanism.
\modify{As shown in Figure~\ref{fig:demo}(a), we manually design an instructive meta-prompt.}
Unlike previous meta-prompts which simply guide LLMs to reflect on the current case, our instructive meta-prompt further directs LLMs to generate exemplars by selecting typical wrong samples and providing detailed solution processes for them.
Thanks to the detailed solution processes within exemplars, LLMs therefore yield more informative feedbacks.
\modify{We then propose Feedback Memory to store all feedbacks and assign a priority score to each of them, as shown in Figure~\ref{fig:demo}(b)}.
During the optimization process, we retrieve a group of feedbacks with the highest priority scores and instruct LLMs to generate a new prompt for 
the feedbacks.
After evaluating the refined prompts, we update the priority scores of the associated feedbacks accordingly, \emph{i.e.,} we increase the score for improved performance and decrease it if no gain.
Consequently, feedbacks with valuable insights will be consistently selected rather than ignored throughout the optimization process.
\modify{As demonstrated in Figure~\ref{fig:demo}(c), we store all exemplars in Exemplar Factory and assign a prior score to each piece.}
At the inference stage, we retrieve a set of exemplars with the highest priority scores, 
and concatenate the exemplars to our refined prompt to further improve the performance. 

We conduct an extensive evaluation on seven tasks to compare ERM with the latest prompt optimization approaches. 
Our results demonstrate substantial improvements over state-of-the-art methods, notably achieving a 10.1 F1 score improvement on the LIAR dataset.
Furthermore, the optimization speed of ERM is roughly twice as fast as ProTeGi~\citep{pryzant2023automatic}.

Our contributions are summarized as follows:

1) We design an instructive meta-prompt, which guides LLMs to select exemplars and therefore yield more informative feedbacks.

2) 
We propose a Feedback Memory to store historical feedbacks by their priority scores, enabling effective retrieval and utilization of feedbacks for prompt optimization.

3) We propose an Exemplar Factory to store and evaluate exemplars. By retrieving exemplars and concatenating them to our refined prompt at the inference stage, we further enhance the performance of LLMs.

4) {We conduct extensive experiments on various tasks and show superior performance of our method to previous state-of-the-arts. Additionally, our optimization steps can be largely reduced, \emph{e.g.}, the steps of our method are approximately half of that in ProTeGi.}

\section{Related Work}

\begin{figure*}[ht]
\centering
  \includegraphics[width=0.98\linewidth]{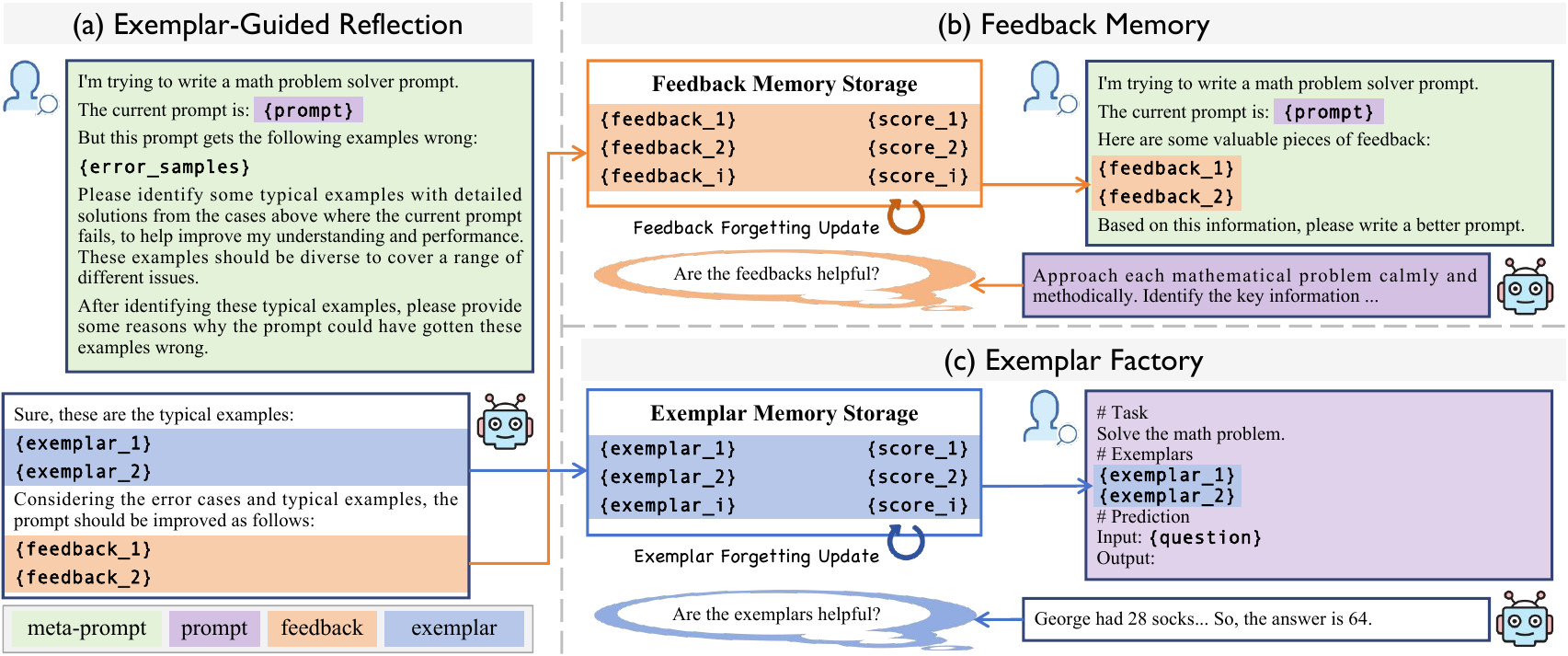}
  \caption{
  Pipeline of ERM.
In wrong prediction samples, the instructive reflective meta-prompt is employed to select exemplars with detailed answer processes, which are subsequently followed by feedback generation. 
The feedbacks are stored in feedback memory storage, and the exemplars are stored in exemplar memory storage. 
These stored feedbacks are periodically retrieved to efficiently guide prompt optimization, with selective forgetting based on their effectiveness in enhancing optimization. 
Additionally, these exemplars are assessed to enhance prediction accuracy.
  }
  \label{fig:pipeline}
\end{figure*}

\subsection{Automatic Prompt Optimization}

Prompt engineering~\cite{zhou2022large} aims to identify suitable prompts as inputs for large language models (LLMs) to perform various tasks.
To minimize human effort, researchers have explored automatic prompt optimization~\cite{lester2021power,shin2020autoprompt,li2021prefix}.
Previous works adopt various strategies for automatic prompt optimization, such as evolutionary-based methods, trajectory-based methods, and feedback-based methods.
Evolutionary-based methods~\citep{guo2023connecting,fernando2023promptbreeder} utilize LLMs to rewrite a set of prompts with evolutionary algorithms~\citep{holland1992genetic,storn1997differential}, which select the best prompts on a validation set to simulate the natural selection process for optimizing prompts.
Trajectory-based methods~\citep{yang2024opro,tang2024unleashing} employ an LLM prompt optimizer to generate new prompts based on historical prompts, scores, or error examples.
Feedback-based methods~\citep{pryzant2023automatic,juneja2024task} use LLMs to summarize feedbacks on 
erroneous cases, 
leveraging the feedbacks to optimize and create new prompts.
In this work, we primarily focus on feedback-based methods, with the aim of writing stronger feedbacks and efficiently utilizing them for optimization.

\subsection{Long-Term Memory Mechanisms}
Existing automatic prompt optimization methods~\citep{pryzant2023automatic,juneja2024task} face challenges in maintaining a robust long-term memory function, limiting their ability to retain and utilize valuable feedbacks for prompt optimization.
MemoryBank~\citep{zhong2024memorybank} solves the challenge of maintaining a robust long-term memory conversation history in previous LLMs~\citep{touvron2023llama,zeng2022glm,taori2023stanford} by introducing a mechanism that enhances their ability to store and recall relevant information over time. 
This approach mimics human memory dynamics through a selective retention strategy inspired by the Ebbinghaus Forgetting Curve~\citep{ebbinghaus2013memory}.
Our work builds on these advancements by using memory storage to implement feedbacks and exemplars in long-term memory.
We implement a forgetting strategy for feedbacks and exemplars that are retrieved but deemed unvaluable, thereby enhancing the efficiency and accuracy of long-term memory retention in prompt optimization.

\section{Method}
\label{sec:method}

In this section, we propose \shortname{}, a novel method designed to achieve efficient and accurate prompt optimization.
As shown in Figure~\ref{fig:pipeline},
\shortname{} is composed of three core components:
(1) \textbf{Exemplar-Guided Reflection}, employing an instructive meta-prompt~(Section~\ref{sec:method-meta-prompt}), guides \promptoptimizer{} to first generate exemplars by identifying typical wrong samples and providing detailed solution processes, followed by generating feedback.
(2) We then propose a \textbf{Feedback Memory}~(Section~\ref{sec:method-feedback-memory}) to store all feedbacks and assign a priority score to each piece of them. 
These feedbacks can then be retrieved and utilized during optimization efficiently. 
After evaluating the refined prompts, we update the priority scores of the associated feedbacks.
(3) Finally, we utilize an \textbf{Exemplar Factory} (Section~\ref{sec:method-examples-memory}) to store and evaluate exemplars, which serve as additional resources during prediction. 
By incorporating the retrieved exemplars into our refined prompt, \taskmodel{} are further guided to achieve improved accuracy.

\subsection{Preliminary}
Given a training set 
\modify{$\mathcal{D}_{train} = \{(q_i,a_i)\}^{n_{t}}_{i=1}$ }
($q_i$ represents the question, $a_i$ is the paired answer, \modify{and $n_{t}$ is the total number of training samples}) and a test set $\mathcal{D}_{test}$ drawn from a specific task, 
along with a score function $s(\cdot)$ for this task, 
we aim to perform the task using a black-box \taskmodel{} $M_{s}$~(\emph{e.g.}, ChatGPT), which combines the prompt $p$ with questions from $\mathcal{D}_{test}$ as input to generate responses. 
These responses are then evaluated by the score function to calculate an average score over $\mathcal{D}_{test}$.
The goal of prompt optimization is to find an optimal prompt $p^*$ drawn from the natural language space that maximizes the expectation of the average score over $\mathcal{D}_{test}$:
\begin{equation}
    p^{*} = \arg\max_{p} \mathbb{E}_{(q_{i}, a_{i})\sim \mathcal{D}_{test}}[s(M_s(q_i; p), a_i)],
\end{equation}
where $p = [p_{I}, p_{R}(q_i)]$ might be composed of two parts: 
one includes the invariant content $p_{I}$, which remains independent of the question and may include task descriptions and general solution steps,
and the other is the variable content $p_{R}(q_i)$ , which is question-specific.
We leverage a more powerful \promptoptimizer{} $M_{e}$ ~(\emph{e.g.}, GPT-4) compared with the \taskmodel{} $M_{s}$ to summarize feedbacks and optimize the prompt. 

Previous work typically divides prompt optimization into three steps: 
prompt initialization, new prompt proposal, and prompt search.

\textbf{1) Prompt initialization.} 
Prompt initialization can be achieved by both manual initialization and induction initialization. 
Following ProTeGi~\citep{pryzant2023automatic}, we initialize the original prompt $p^{0}$ manually.

\textbf{2) New prompt proposal.}
\modify{Commonly, previous methods use \taskmodel{} $M_s$ to evaluate on a subset of $\mathcal{D}_{train}$, and then use  \promptoptimizer{} $M_e$ to summarize errors from $n_{b}$ wrong samples $\mathcal{B}=\{(q_i, a_i)\}_{i=1}^{n_{b}}$,  where the response of \taskmodel{} $M_s(q_i, p^{t})$ is different from $a_i$.} 
Feedbacks is then generated as $\mathcal{F}^{t} = M_{e}(p^{t}, \mathcal{B}; p^{meta}_{ref})$ with $p^{meta}_{ref}$ serving as the meta-prompt that guides the \promptoptimizer{} in generating feedback.
The \promptoptimizer{} then optimizes and refines the prompt $p^{t}$ based on the feedbacks to obtain refined prompts $p^{t+1}=M_{e}(p^{t}, \mathcal{B}, f^{t}; p^{meta}_{opt})$, where $f^{t} \in \mathcal{F}^{t}$, and $p^{meta}_{opt}$ is the meta-prompt guiding the \promptoptimizer{} to propose refined prompt.

\textbf{3) Prompt search.} 
Following ProTeGi, we employ a beam search strategy to further select the refined prompts. 
Among several candidate prompts $\mathcal{P}^{t+1}$, we select $k$ prompts which perform best on the validation set, which is the subset of the training set.
These $k$ prompts are then used for the next optimization step.

\subsection{Exemplar-Guided Reflection}
\label{sec:method-meta-prompt}
To encourage the \promptoptimizer{} generate more informative feedbacks, we propose an Exemplar-Guided Reflection in Figure~\ref{fig:pipeline}(a), which utilizes an instructive meta-prompt to select typical wrong samples with detailed solution processes as exemplars and generate feedbacks for them.
Detailedly, we first utilize the instructive meta-prompt $p^{meta}_{ref*}$, which guides the \promptoptimizer{} to select \modify{$n_{e}$} diverse and significantly representative wrong samples from the wrong samples $\mathcal{B}$ as exemplars $\mathcal{E}^{t}$ and provide detailed solution processes for them:
\begin{align}
    \mathcal{E}^{t} &= M_e(p^{t}, \mathcal{B}; p^{meta}_{ref*}),
\end{align}
\modify{
where 
$\mathcal{E}^{t} = \{e_i\}_{i=1}^{n_{e}} = \{(q_i, a_i, \mathrm{cot}_i)\}_{i=1}^{n_{e}}$ 
is a set of exemplars $e_i$ with detailed solution processes $\mathrm{cot}_i$.
}
\modify{Then, the \promptoptimizer{} generates $n_{f}$ feedbacks 
$\mathcal{F}^{t}=\{f^{t}_{i}\}_{i=1}^{n_{f}}$}
, which offer insights on example predictions and suggestions on modification of the prompt:
\begin{align}
    \mathcal{F}^{t} &= M_e(p^{t}, \mathcal{B}, \mathcal{E}^{t}; p^{meta}_{ref*}),
\end{align}
Based on the wrong samples $\mathcal{B}$ and each item in the generated feedbacks $f^{t} \in \mathcal{F}^{t}$, the model finally produce a refined prompt $p^{t+1}$ for each feedback: 
\begin{equation}
    p^{t+1} = M_e(p^{t}, \mathcal{B}, f^{t}; p^{meta}_{opt}).
\end{equation}

\subsection{Feedback Memory}
\label{sec:method-feedback-memory}

Aiming to accelerate the convergence of prompt optimization process, we propose a Feedback Memory in Figure~\ref{fig:pipeline}(b).
We store the feedbacks with priority scores via a long-term memory mechanism and retrieve them efficiently for optimization.
By evaluating the generated prompts, we selectively forget the feedbacks to ensure that all stored feedbacks remain beneficial for prompt optimization.

\noindent
\textbf{Feedback Memory Storage}
In Feedback Memory, we store the valuable feedbacks during the optimization process and assign a priority score to each piece of them, which serves as a basic foundation for Feedback Forgetting Updating. 
To effectively store useful feedbacks and prevent adverse impacts on prompt optimization, we employ a feedback filtering strategy:
(1) We evaluate the refined prompts generated based on the feedbacks, and only store the informative feedbacks whose corresponding prompts bring improvements on the validation set.
Such strategy ensures that only valuable feedbacks are stored and retrieved.
(2) Additionally, we employ the BGE-M3 model~\citep{chen2024bge} to calculate the semantic similarity between newly generated feedbacks and the stored ones.
We ignore the feedbacks of high similarity with the previous ones to avoid redundant information.

\noindent
\textbf{Feedback Retrieval}
During the optimization process, we periodically select historical feedbacks from the memory based on their priority scores.
Specifically, we calculate the selection probability for each feedback as follows:
\begin{equation}
    P_f = \text{softmax}\left(\left\{e^\frac{s_p(f_i)}{\tau_f}\right\}_{i=1}^{|\tilde{\mathcal{F}}|}\right),
\end{equation}
where ${\tau_f}$ is the temperature, controlling the tendency to select high-scoring feedbacks, and $\tilde{\mathcal{F}}$ denotes all feedbacks stored in the memory. 
\modify{We then randomly select $n_{\hat{f}}$ feedbacks according to their selection probabilities:}
\begin{equation}
   \modify{\hat{\mathcal{F}} = \{f_{i}\}_{i=1}^{n_{\hat{f}}} = \text{sample}(\tilde{\mathcal{F}}, P_f).}
\end{equation}

\noindent
\textbf{Feedback Forgetting Updating}
The selected feedbacks $\hat{\mathcal{F}}$ guide \promptoptimizer{} generate new prompts $p^{t+1} = M_e(p^t, \mathcal{B}, \hat{\mathcal{F}}; p^{meta}_{opt*})$, where $p^{meta}_{opt*}$ is the meta-prompt that efficiently utilizes the feedback group to generate a refined prompt.
We then update their priority scores by evaluating the generated prompt: we increase the priority score if the performance is improved but decrease it if no gain.
\begin{equation}
    s^t_p(f) = (1 - \beta)s_p(f)^{t-1} + \beta\mathbb{I}(f),
\end{equation}
where $\mathbb{I}(f)$ represents whether sufficient performance gain is achieved and $\beta$ is a hyper-parameter to control the speed of updating.
Besides, the feedback will be removed from the storage once its priority score falls below a certain threshold \modify{$\theta$}:
\begin{equation} 
\modify{\tilde{\mathcal{F}}^{t} = \{ f \mid f \in \tilde{\mathcal{F}}^{t-1} , s^t_p(f) \geq \theta \} .}
\end{equation}
With such Forgetting Updating mechanism, we ensure that the most valuable feedbacks are continuously utilized, which efficiently accelerate the convergence of our optimization process.

\subsection{Exemplar Factory}
\label{sec:method-examples-memory}

As shown in Figure~\ref{fig:pipeline}(c),
we store the exemplars along with a priority score to each piece of them, similar to that in Feedback Memory. 
These exemplars are stored in memory and retrieved for prediction, allowing us to assess their impact on the task. 
We selectively forget exemplars, ensuring that the valuable ones will be retrieved to enhance the prediction performance.

\noindent\textbf{Exemplar Memory Storage}
The exemplar memory storage retains valuable exemplars. 
We introduce an exemplar filtering strategy to ensure stored exemplars benefit prediction:
(1) We verify that the detailed solution process of the exemplar generated by \promptoptimizer{} matches to the ground truth label.
(2) When a new generated exemplar is identical to the stored ones, we replace the stored exemplars with probability $p$ and reject the new exemplar with probability $1-p$ to avoid redundant storage.

\noindent\textbf{Exemplar Retrieval}
Each exemplar $e_i$ is assigned a priority score $s_p(e_i)$. During the prompt optimization process for question $q_j$, we calculate the selection probability for each exemplar as follows:
\begin{equation}
    P^r_e = \text{softmax}\left( \left\{
       e^{\frac{s_p(e_i) \cdot s_s^j(e_i)}{\tau_e}}
    \right\}_{i=1}^{|\tilde{\mathcal{E}}|}  \right),
\end{equation}
where $s_p(e_i)$ is the priority score of exemplar $e_i$ and $s_s^j(e_i)$ is its semantic similarity to the question $q_j$, $\tilde{\mathcal{E}}$ represents the stored exemplars, and $\tau_{e}$ is the temperature. 
We then randomly sample five exemplars as variable content $p_{R}(q_j)$ of prompt.
During the inference stage, we select the five exemplars with the highest $s_p(e_i) \cdot s_s^j(e_i)$ as variable content $p_{R}(q_j)$ of prompt for more accurate predictions.

\noindent\textbf{Exemplar Forgetting Updating}
We adjust the priority scores of exemplars based on whether incorporating them as the variable content $p_{R}(q_j)$ in the prompt leads to improvements. 
Exemplars with low priority scores are promptly removed to ensure that only valuable ones are stored.

\begin{table*}[!t]
\setlength{\tabcolsep}{2.2pt}
\centering
\scalebox{0.84}{
\begin{tabular}{lccccccc}
\toprule
\multirow{2}{*}{\raisebox{-1.9ex}[0pt][0pt]{Method}} & \multicolumn{4}{c}{True / False} & \multicolumn{2}{c}{Generative} & Multiple-choice \\ \cmidrule(lr){2-5} \cmidrule(lr){6-7} \cmidrule(lr){8-8}
 & \makecell[c]{\,LIAR\,\\{\small (F1)}} & \makecell[c]{\,BBH\,\\{\small (F1)}}  & \makecell[c]{\,ETHOS\,\\{\small (F1)}} & \makecell[c]{ArSarcasm\\{\small (F1)}} & \makecell[c]{WebNLG\\{\small (Rouge-L)}} & \makecell[c]{GSM8K\\{\small (Acc.)}} & \makecell[c]{\,WSC\,\\{\small (Acc.)}} \\ 
\midrule
Empty        & 46.4 & 69.4 & 93.0  & 83.7      & 49.4   & 89.0  &  77.3\\
CoT~\citep{kojima2022large} & 46.0 & \underline{81.9} & 84.5  & 83.7      & 49.3   & 89.0  &  81.3  \\
\midrule
APE~\citep{zhou2022large} & 47.7 & 72.9 & 94.0  & 83.8      & 51.3   & 91.3  & 79.3  \\
ProTeGi~\citep{pryzant2023automatic} & \underline{58.5} & 73.6 & \underline{96.5}  & 84.1      & \underline{55.7}   & 91.0  &80.0\\
OPRO~\citep{yang2024opro} & 47.9 & 75.7 & 93.5  & \underline{84.5}      & 51.9   & 90.7  &    83.3 \\
Promptbreeder~\citep{fernando2023promptbreeder} & 47.1 & 74.3 & 94.5  & 83.8      & 51.0   & \underline{91.7}  &  80.0 \\
EvoPrompt~\citep{guo2023connecting} & 47.9 & 75.0 & 93.0  & 83.8      & 50.2   & 90.7  & 78.8  \\
GPO~\citep{tang2024unleashing} & 54.7 & 70.8 & 94.0  & 83.6      & 51.8   & 90.3  & \underline{84.0} \\
\midrule
\shortname{} & \textbf{68.6} & \textbf{86.1} & \textbf{98.0}  & \textbf{85.1}      & \textbf{59.6}   & \textbf{93.3}  & \textbf{86.0} \\
$\Delta$ & \green{+10.1} & \green{+4.2} & \green{+1.5}  & \green{+0.6}      & \green{+3.9}   & \green{+1.6}  & \green{+2.0} \\
\bottomrule
\end{tabular}
}
\caption{Comparisons of our method with existing LLM-based prompt optimizers under zero-shot setting.}
\label{tab:main_result}
\end{table*}

\begin{table*}[!t]
\centering
\scalebox{0.79}{
\begin{tabular}{P{2.5cm}P{14.5cm}C{1.5cm}}
\toprule
\makecell[c]{Method} & \makecell[c]{Prompt} & \makecell[c]{Rouge-L} \\
\midrule
Empty & Write the following triples as fluent English text. & 49.4 \\
\hdashline \noalign{\vskip 0.5ex}
ProTeGi & You are given a set of triples that need to be converted into coherent and fluent English sentences. Each triple consists of a subject, predicate, and object. Your task is to accurately convey the information from these triples into well-formed sentences. Ensure the sentences are complete, grammatically correct, and clearly express the relationships provided in the triples. & 55.7 \\
\hdashline \noalign{\vskip 0.5ex}
OPRO & Convert the following sets of triples into coherent, natural, and fluent English sentences. & 51.9 \\
\hdashline \noalign{\vskip 0.5ex}
PromptBreeder & Transform these triples into smooth and stylish English sentences, and make them shine! & 50.9 \\
\hdashline \noalign{\vskip 0.5ex}
EvoPrompt & Turn the provided triples into smooth, flowing English sentences that will impress everyone! & 50.2 \\
\hdashline \noalign{\vskip 0.5ex}
GPO & Rewrite these triples into fluent and natural English sentences. & 51.8 \\
\hdashline \noalign{\vskip 0.5ex}
\shortname{} & Convert the following triples into coherent and fluent English sentences. Ensure that all relationships and attributes are accurately conveyed. When multiple associations or attributes are involved, break down the information into smaller, logical sentences to maintain clarity. & 59.6 \\
\bottomrule
\end{tabular}
}
\caption{Prompts optimized by different methods on the WebNLG dataset.
}
\label{tab:webnlg_instructions}
\end{table*}

\section{Experiments}

\noindent
\textbf{Datasets.}
We perform evaluation on 7 standard datasets : 
WSC~\citep{levesque2012winograd},
Ethos~\citep{mollas2022ethos},
ArSarcasm~\citep{farha2020arabic},
Liar~\citep{wang2017liar},
BBH-navigate~\citep{suzgun2022challenging},
GSM8k~\citep{cobbe2021training},
WebNLG~\citep{gardent2017creating}.
Among these, ArSarcasm, Ethos, and Liar, and BBH-navigate contain true/false questions, 
WSC contains multiple-choice questions, 
GSM8K contains questions with integer answers, 
and WebNLG contains questions requiring natural language generation.

\noindent
\textbf{Baselines.}
We compare several representative methods, including existing LLM-based prompt optimizers: APE~\citep{zhou2022large}, \modify{ProTeGi}~\citep{pryzant2023automatic}, OPRO~\citep{yang2024opro}, Promptbreeder~\citep{fernando2023promptbreeder}, EvoPrompt~\citep{guo2023connecting}, and GPO~\citep{tang2024unleashing}.
In addition, we consider the baseline using manually written simple prompts (``Manual''), which we provide in the appendix, and the instruction ``Let's think step by step.'' from chain-of-thought prompting (``CoT'')~\citep{kojima2022large} for performance comparison.

\noindent
\textbf{Evaluation Metrics.} 
We report the F1 score on Ethos, ArSarcasm, Liar and BBH-navigate following~\citep{pryzant2023automatic}, accuracy on WSC and GSM8k following~\citep{tang2024unleashing, juneja2024task} and ROUGE-L on WebNLG following~\citep{tang2024unleashing}.

\noindent
\textbf{Implementation Details.} 
For the \taskmodel{}, we use Doubao-Pro~\citep{doubao}. For the \promptoptimizer{}, we use GPT-4o~\citep{gpt4o}.
We repeat all the experiments three times and report the average of the results.
Other details are presented in appendix.

\begin{figure*}[!t]
  \includegraphics[width=0.98\linewidth]{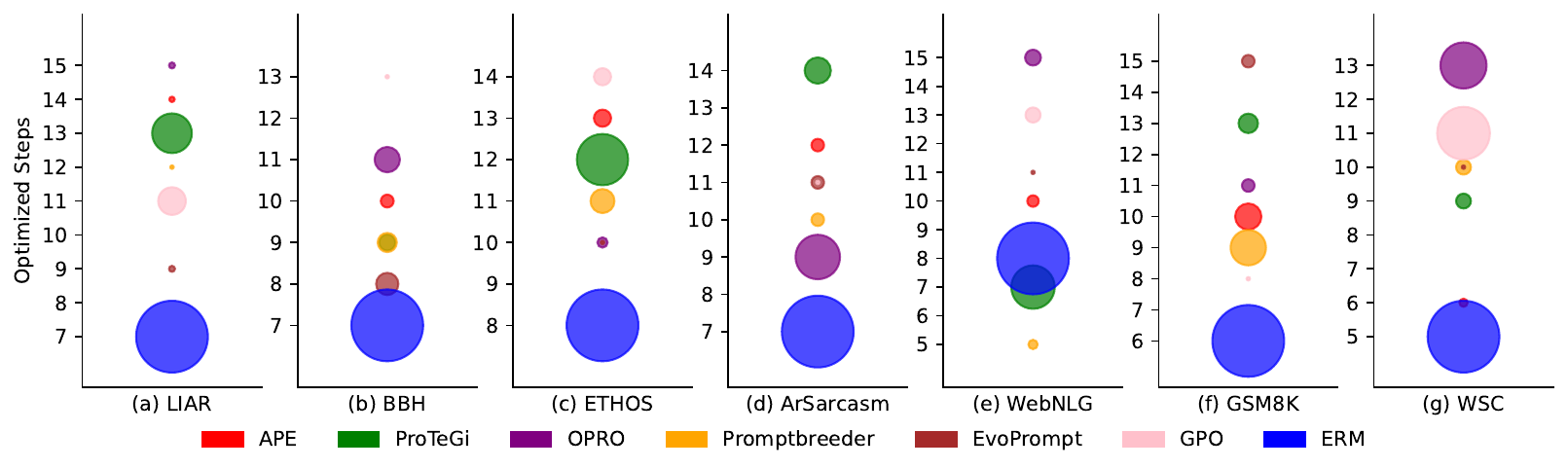}
  \caption{The efficiency of our approach \shortname{}.
  The size of the circles represents performance, with larger circles indicating better performance. The vertical axis shows the optimization steps needed for different methods to achieve peak performance across datasets.
  }
  \label{fig:efficiency}
\end{figure*}

\subsection{Main Results}

\begin{table*}[!t]
\setlength{\tabcolsep}{2.2pt}
\centering
\scalebox{0.84}{
\begin{tabular}{lccccccc}
\toprule
\multirow{2}{*}{\makecell[c]{\raisebox{-1.9ex}[0pt][0pt]{Method}}} & \multicolumn{4}{c}{True / False} & \multicolumn{2}{c}{Generative} & Multiple-choice \\ \cmidrule(lr){2-5} \cmidrule(lr){6-7} \cmidrule(lr){8-8}
 & \makecell[c]{\,LIAR\,\\{\small (F1)}} & \makecell[c]{\,BBH\,\\{\small (F1)}}  & \makecell[c]{\,ETHOS\,\\{\small (F1)}} & \makecell[c]{ArSarcasm\\{\small (F1)}} & \makecell[c]{WebNLG\\{\small (Rouge-L)}} & \makecell[c]{GSM8K\\{\small (Acc.)}} & \makecell[c]{\,WSC\,\\{\small (Acc.)}} \\ 
\midrule
APE~\citep{zhou2022large}           & 51.2 & 74.3 & 93.2  & 84.3      & 53.1   & \underline{91.8}  & 80.3 \\
ProTeGi~\citep{pryzant2023automatic} & \underline{60.3} & 73.6 & \underline{97.0}  & 84.1      & \underline{56.3}   & 91.0  & 81.0 \\
OPRO~\citep{yang2024opro} & 52.1 & 75.0 & 94.8  & \underline{84.7}      & 52.4   & 90.8  & \underline{85.0} \\
Promptbreeder~\citep{fernando2023promptbreeder}  & 51.8 & 75.7 & 95.7  & 84.5      & 52.7   & 91.7  & 81.5 \\
EvoPrompt~\citep{guo2023connecting}& 52.3 & \underline{76.4} & 94.3  & 83.9      & 51.8   & 90.9  & 80.4 \\
GPO~\citep{tang2024unleashing} & 56.6 & 75.0 & 95.5  & 83.8      & 53.4   & 90.5  & 84.9 \\
\midrule
\shortname{} & \textbf{68.6} & \textbf{86.1} & \textbf{98.0}  & \textbf{85.1}      & \textbf{59.6}   & \textbf{93.3}  & \textbf{86.0} \\ 
\bottomrule
\end{tabular}
}
\caption{Comparisons of our method with existing LLM-based prompt optimizers under few-shot setting.}
\label{tab:main_result_few_shot}
\end{table*}

\begin{table*}[!t]
\centering
\setlength{\tabcolsep}{2.2pt}
\scalebox{0.785}{
\begin{tabular}{cccccccccc}
\toprule
\makecell[c]{Exemplar-Guided\\Reflection} & Feedback Memory & Exemplar Factory & \makecell[c]{\,LIAR\,\\{\small (F1)}} & \makecell[c]{\,BBH\,\\{\small (F1)}} & \makecell[c]{\,ETHOS\,\\{\small (F1)}} & \makecell[c]{ArSarcasm\\{\small (F1)}} & \makecell[c]{WebNLG\\{\small (Rouge-L)}} & \makecell[c]{GSM8K\\{\small (Acc.)}} & \makecell[c]{\,WSC\,\\{\small (Acc.)}} \\
\midrule
& & & 58.5 & 73.6 & 96.5  & 84.1 & 55.7 & 91.0  & 80.0 \\
\ding{51} & & & 62.9 & 75.7 & 97.0  & 84.2 & 56.9 & 92.7  & 82.0 \\
\ding{51} & & \ding{51} & 67.2 & 84.7 & 97.0  & 84.9 & 58.6 & 93.0 & 84.0 \\
\ding{51} & \ding{51} & & 66.6 & 82.6 & 97.5  & 84.8 & 58.8 & 93.0  & 85.0 \\
\ding{51} & \ding{51} & \ding{51} & 68.6 & 86.1 & 98.0  & 85.1 & 59.6 & 93.3  & 86.0 \\
\bottomrule
\end{tabular}
}
\caption{Effect of each component in our method.}
\label{tab:effect_of_each_component}
\end{table*}

\noindent
\textbf{Comparison under Zero-shot Setting.}
Table~\ref{tab:main_result} presents the results of different methods for prompt optimization across true/false questions, generative questions, and multiple-choice questions.

For true/false questions, our method demonstrates a significant improvement over previous works. 
Specifically, our method outperforms trajectory-based methods (OPRO and GPO) by 13.9. 
Trajectory-based methods utilize an LLM prompt optimizer to generate new prompts based on historical prompts, scores, or error examples, but may struggle to identify ``better prompts'', limiting their performance.
Our method also outperforms ProTeGi (feedback-based method) by 10.1, which can be attributed to our exemplar-guided reflection, feedback memory and example factory. 

For generative questions and multiple-choice questions, our method also significantly outperforms previous methods. 
Specifically, on the WebNLG dataset, our approach surpasses previous methods by 3.9 in Rouge-L score. 
Table~\ref{tab:webnlg_instructions} visualizes the optimized prompts on the WebNLG dataset, demonstrating that our method's optimized prompts are more effective at capturing the critical information needed to enhance task performance.
The exemplar factory boosts the F1 score by 3.7 on the LIAR dataset, while the feedback memory provides an improvement of 2.0.

\noindent
\textbf{Efficiency of Our Method.}
Our approach introduces a memory mechanism to efficiently store and utilize feedbacks. 
We show the optimization steps needed for different methods to achieve peak performance across datasets in Figure~\ref{fig:efficiency}, which highlights the superior efficiency of our method.
Specifically, according to Figure~\ref{fig:efficiency}(a), on the LIAR dataset, our method reaches an F1 score of 68.6 by the 7th step, while ProTeGi only achieves 58.5 by the 13th step, demonstrating that our method nearly doubles the optimization speed.

\noindent
\textbf{Comparison under Few-shot Setting.}
Table~\ref{tab:main_result_few_shot} presents a comparison between our method and others under few-shot settings. 
For each approach, we dynamically select five relevant examples through k-nearest neighbors (kNN) clustering in the embedding space. 
According to the results, ERM consistently outperforms the previous methods. 
Notably, on the LIAR dataset, our approach achieves an 8.3 F1 score improvement over previous methods, demonstrating the effectiveness of selecting valuable wrong examples as exemplars and equipping them with chain-of-thought-like solution processes.

\subsection{Ablation Study}

\noindent \textbf{Effect of Each Component.} 
In Table~\ref{tab:effect_of_each_component}, we conduct experiments to verify the effectiveness of each key component in our method. 

We adopt a strategy which dentify exemplars, contemplate the corresponding chain of thought and then complete feedbacks, and observe that ERM improves the F1 score by 4.4 on the LIAR dataset compared with the approaches without the instructive meta-prompt, which validates the effectiveness of the instructive meta-prompt.
Additionally, the introduction of the memory mechanism for feedback memory and exemplar factory brought a further 5.7 improvement on the LIAR dataset, confirming the effectiveness of the memory mechanism.

\begin{table}[!t]
\centering
\setlength{\tabcolsep}{4.8pt}
\scalebox{0.75}{
\begin{tabular}{cccccc} 
\toprule
Retrieval & \makecell[c]{Exemplar\\Filtering} & \makecell[c]{Selective\\Forget.} & \makecell[c]{\,LIAR\,\\{\small (F1)}} & \makecell[c]{\,BBH\,\\{\small (F1)}} & \makecell[c]{WebNLG\\{\small (Rouge-L)}} \\
\midrule
 &  & & 62.9 & 75.7 & 56.9    \\
\ding{51}  & & & 62.3 & 75.0 & 57.0    \\
\ding{51}  & \ding{51} & & 65.7 & 81.3 & 58.4    \\
\ding{51} & \ding{51} & \ding{51} & 66.6 & 82.6 & 58.8    \\
\bottomrule
\end{tabular}
}
\caption{Effect of each component in Exemplar Factory.}
\label{tab:memory_anchor_examples}
\end{table}

\noindent
\textbf{Effect of Exemplar Factory.}
As shown in Table~\ref{tab:memory_anchor_examples}, incorporating exemplar filtering when storing exemplars does not enhance performance. This is because the behavior of the \promptoptimizer{} is unpredictable and may generate incorrect or unconventional questions.
Retrieving such examples does not enhance prediction performance. 
However, filtering out erroneously generated exemplars and redundant ones already in storage resulted in a 3.4 improvement, highlighting the importance of exemplar filtering.
The introduction of a selective forgetting further improved the F1 score by 0.9 on the LIAR dataset, as it removes exemplars that do not aid in prediction, thereby enhancing performance.

\begin{table}[!t]
\setlength{\tabcolsep}{4.8pt}
\scalebox{0.75}{
\begin{tabular}{cccccc}
\toprule
Retrieval & \makecell[c]{Feedback\\Filtering} & \makecell[c]{Selective\\Forget.} & \makecell[c]{\,LIAR\,\\{\small (F1)}} & \makecell[c]{\,BBH\,\\{\small (F1)}} & \makecell[c]{WebNLG\\{\small (Rouge-L)}} \\
 \midrule
 & & & 66.6 & 82.6 & 58.8    \\
\ding{51} & & & 66.4 & 81.9 & 58.8    \\
\ding{51} & \ding{51} & & 67.5 & 82.6 & 59.2    \\
\ding{51} & \ding{51} & \ding{51} & 68.6 & 86.1 & 59.6    \\
\bottomrule
\end{tabular}
}
\caption{Effect of each component in Feedback Memory.}
\label{tab:memory_feedback_xx}
\end{table}

\noindent \textbf{Effect of Feedback Memory.}
As shown in Table~\ref{tab:memory_feedback_xx}, directly storing feedbacks for periodic optimization without the feedback filtering strategy does not improve performance. 
Introducing the filtering strategy increased the F1 score on the LIAR dataset by 0.9 compared to not using stored feedbacks.
Additionally, incorporating selective forgetting, which discards suboptimal feedback promptly, further enhanced the F1 score by an additional 0.9.

\section{Conclusion}

In this paper, we introduce \textbf{E}xemplar-Guided \textbf{R}eflection with \textbf{M}emory mechanism~(\shortname{}), a novel approach to achieve efficient and accurate prompt optimization.
Using a instructive reflection meta-prompt, \shortname{} instructs LLMs to select exemplars with detailed solution processes and generate stronger feedback. 
We then propose Feedback Memory mechanism to efficiently exploit potentially valuable feedback. 
Additionally, Exemplar Factory is introduced to 
further enhance the accuracy of prediction by pre-assessing the impact on the task.
\shortname{} refines prompts authored by human experts and outperforms established automatic prompt engineering baselines across various scenarios, with optimization steps approximately half of that in previous work.

\section{Limitations}

In this work, we effectively utilize feedbacks and exemplars using a long-term memory mechanism. 
However, in real-world applications, we encounter additional challenges: some questions continue to be incorrectly answered during the optimization process, and prompt optimization doesn’t always align with human expectations. 
When the model struggles to optimize, introducing human intervention might aid in enhancing prompt optimization.
This paper lacks exploration on how humans could assist in the optimization process. 
For instance, with persistent incorrect answers, human input could offer crafted solutions, helping the expert model generate improved feedback. 
Additionally, due to computational and budget constraints, our experiments are limited to representative tasks.


\bibliography{custom}


\appendix

\begin{table*}[h]
\setlength{\tabcolsep}{8pt}
\centering
\scalebox{0.75}{
\begin{tabular}{llll}
\toprule
Dataset Name & Task & Train \& Dev & Test \\
\midrule
LIAR~\citep{wang2017liar} & True/False & 3681         & 461  \\
BBH-Navigate~\citep{suzgun2022challenging} & True/False & 96           & 144  \\
ETHOS~\citep{mollas2022ethos} & True/False & 440          & 200  \\
ArSarcasm~\citep{farha2020arabic} & True/False & 8437         & 2110 \\
WebNLG~\citep{gardent2017creating} & Language Generation & 200          & 300  \\
GSM8K~\citep{cobbe2021training} & Integer Generation & 200          & 300  \\
WSC~\citep{levesque2012winograd} & Multiple-Choice  & 100          & 150  \\
\bottomrule
\end{tabular}}
\caption{Dataset sizes and data splits.}
\label{tab:dataset_split}
\end{table*}

\begin{table*}[h]
\setlength{\tabcolsep}{2pt}
\centering
\scalebox{0.75}{
\begin{tabular}{lll}
\toprule
Dataset                     & License & Source \\
\midrule
LIAR~\citep{wang2017liar} & Unknown & \url{https://www.cs.ucsb.edu/~cwilliam/data/liar_dataset.zip} \\
BIG-bench Hard~\citep{suzgun2022challenging} & Apache-2.0 & \url{https://github.com/google/BIG-bench} (original) \\
& & \url{https://github.com/suzgunmirac/BIG-Bench-Hard} (reformatted) \\
ETHOS~\citep{mollas2022ethos} & GNU GPLv3 & \url{https://huggingface.co/datasets/iamollas/ethos}  \\
ArSarcasm~\citep{farha2020arabic} & MIT & \url{https://github.com/iabufarha/ArSarcasm}  \\
WebNLG~\citep{gardent2017creating} & CC BY 4.0 & \url{https://github.com/fuzihaofzh/webnlg-dataset}  \\
GSM8K~\citep{cobbe2021training} & MIT & \url{https://github.com/openai/grade-school-math}  \\
WSC~\citep{levesque2012winograd} & CC BY 4.0 & \url{https://huggingface.co/datasets/ErnestSDavis/winograd_wsc}  \\
\bottomrule 
\end{tabular}
}
\caption{License and Source of the datasets used in this study.}
\label{tab:dataset_license}
\end{table*}

\section{Additional Details for the Setup}

\subsection{Tasks and Data Details}

We present a summary of the dataset sizes and data split information in Table~\ref{tab:dataset_split}. Table~\ref{tab:dataset_license} provides details on the sources and licensing information of the datasets. To the best of our knowledge, our usage of these datasets aligns with their intended purposes, and the data we utilize do not contain any personal or sensitive information.

\noindent
\textbf{LIAR}~\citep{wang2017liar} is an English fake news detection corpus comprising 4,000 statements, each accompanied by context and lie labels. 
For our experiments, we adopt the same dataset split as ProTeGi~\citep{pryzant2023automatic}, utilizing 3,681 instances for training and 461 instances for testing.

\noindent
\textbf{BIG-bench Hard} dataset~\citep{suzgun2022challenging} is a subset of the BIG Bench dataset~\cite{srivastava2022beyond}, comprising 23 tasks that present significant challenges for current language models. For our experiments, we select the navigation task, which requires determining whether an agent, following a series of navigation steps, returns to its initial starting point. Consistent with the dataset split used by GPO~\citep{tang2024unleashing}, we employ 96 instances for training and 144 instances for testing.

\noindent
\textbf{ETHOS}~\citep{mollas2022ethos} is an English hate speech detection dataset consisting of 997 online comments, each annotated with hate speech labels. In accordance with previous research, we utilize the same dataset split, employing 440 instances for training and 200 instances for testing.

\noindent
\textbf{ArSarcasm} dataset~\citep{farha2020arabic} is an Arabic sarcasm detection corpus containing 10,000 online comments, each labeled for sarcasm. We utilize the original dataset split, with 8,437 instances designated for training and 2,110 instances for testing.

\noindent
\textbf{WebNLG} corpus consists of sets of triplets that describe facts—entities and the relations between them—paired with their corresponding expressions in natural language text. Following the dataset split used by GPO~\citep{tang2024unleashing}, we utilize 200 instances for training and 300 instances for testing.

\noindent
\textbf{GSM8K}~\citep{cobbe2021training} comprises 8.5K high-quality linguistically diverse grade school math word problems, crafted by human problem writers. 
Following the dataset split used by GPO~\citep{tang2024unleashing}, we utilize 200 instances for training and 300 instances for testing.

\noindent
\textbf{WSC} was introduced both as an alternative to the Turing Test and as a measure of a system’s ability to perform commonsense reasoning. 
Following the approach used by GPO~\citep{tang2024unleashing}, we sample 100 examples for the training set and 150 for the test set.

\subsection{Implementation Details}
We select Doubao-pro~\citep{doubao} as the \taskmodel{} and set its temperature to 0, ensuring deterministic outputs following the GPO~\citep{tang2024unleashing} and AgentInstruct~\citep{crispino2023agent}.
For the \promptoptimizer{}, we utilize gpt-4o-2024-05-13, the underlying model of GPT-4o~\citep{gpt4o}. Its temperature is set to 1.0 to promote diverse generation.
The initial prompts for different tasks can be found in Section~\ref{sec:addition_result}.
In each step, the optimizer generates 8 candidate task prompts. 
Following GPO~\citep{tang2024unleashing} and OPRO~\citep{yang2024opro}, the best-performing one is selected as the task prompt for the next iteration.
All experiments are conducted three times, and we report the average results.

\section{More Related Work}

Prompt engineering aims to identify suitable prompts as inputs for large language models (LLMs) to perform various tasks. 
To reduce human effort, researchers have explored automatic prompt optimization~\cite{lester2021power,shin2020autoprompt,li2021prefix}. 

Continuous approaches~\cite{lester2021power,shin2020autoprompt,li2021prefix} optimize within the embedding space of LLMs and update based on backpropagating gradients. 
Prefix tuning~\citep{li2021prefix} introduces new learnable tokens that can be considered as prompts in continuous space, which are learned for specific tasks. 
However, since these tokens are defined in continuous space, they are not easily interpretable, and these methods require access to model weights, making them unsuitable for use with closed-source LLMs like ChatGPT~\citep{chatgpt}.

Discrete methods~\citep{deng2022rlprompt,zhang2022tempera} directly optimize natural language prompts. 
Several strategies have been developed for this purpose. 
Some approaches~\citep{pryzant2023automatic,juneja2024task} optimize prompts based on error feedback, while others~\citep{yang2024opro,tang2024unleashing} utilize multiple prompts and their respective scores to enable the model to identify superior prompts. 
Additionally, certain methods~\citep{guo2023connecting,fernando2023promptbreeder,li2023spell} employ genetic algorithms to rewrite prompts through processes of variation and natural selection. 
Furthermore, some methods~\citep{ye2023prompt,ma2024large} enhance the controllability of feedback generation and prompt optimization by modifying meta-prompts. 
To improve the accuracy of error summaries, some works~\citep{juneja2024task,austin2024grad} cluster similar erroneous samples instead of using randomly selected ones.

\modify{
Recently, automatic prompt optimization has also been explored in the context of multi-step tasks~\citep{chen2024prompt} and multi-modality tasks~\citep{zhu2024vislinginstruct,fan2023towards,hao2024optimizing,um2024minorityprompt,mo2024dynamic,wu2024universal,li2024promptist}.
PROMST~\citep{chen2024prompt} optimizes multi-step tasks by introducing human-designed feedbacks and a score prediction model. 
VisLingInstruct~\citep{zhu2024vislinginstruct} autonomously evaluates and optimizes instructional texts through in-context learning, improving the synergy between visual perception and linguistic expression in multi-modal language models. 
Liu~\citep{liu2024language} enables the language model to rewrite refined prompts based on the scores of historical prompts on image classification datasets.
OPT2I~\citep{manas2024improving} uses a method similar to OPRO, where the language model rewrites better prompts based on the history of prompts and their generated image consistency objective scores.
}


\section{More Experiments}

\noindent
\modify{\textbf{Total Consumed Tokens Comparison.}
As shown in Table~\ref{tab:cpr_consumed_token}, we compare the total number of tokens consumed for different methods in the LIAR dataset.
Our method belongs to the feedback-based prompt optimization category. 
Compared with ProTeGi, our method can efficiently exploit feedbacks with the Feedback Factory, allowing us to optimize with fewer steps and significantly reducing the total number of consumed tokens.
Non-feedback-based methods do not require feedback, so the average number of tokens per step is about half that in feedback-based methods. 
However, since the Feedback Factory can efficiently use feedbacks to reduce optimization steps significantly, the total number of consumed tokens of our method is comparable to those non-feedback-based methods.
}

\begin{table}[!t]
\setlength{\tabcolsep}{4.1pt}
\centering
\scalebox{0.80}{
\begin{tabular}{lccccc}
\toprule
Method         & \gray{APE}   & \gray{OPRO} & \gray{GPO} & ProTeGi & ERM  \\
\midrule
Feedback-based & \gray{\ding{56}} & \gray{\ding{56}} & \gray{\ding{56}} & \ding{51} & \ding{51} \\
F1            & \gray{47.7}    & \gray{47.9} & \gray{54.7} & 58.5 & 68.6 \\
Total. Tokens & \gray{7415} &	\gray{8652}	& \gray{7360}	 &	12836& 7503 \\
\bottomrule
\end{tabular}
}
\caption{
\modify{Total consumed tokens comparison of our method with existing LLM-based prompt optimizers.}
}
\label{tab:cpr_consumed_token}
\end{table}

\noindent
\textbf{Effect of Exemplars' Solutions.}
As shown in Table~\ref{tab:cpr_few_shot}, we compared direct retrieval for prediction on the training set and found that using exemplars yields better results. 
This is because 
(1) the Exemplar Factory pre-assesses exemplars for their effectiveness on the task, filtering useful ones, 
and (2) the \promptoptimizer{} crafts chain-of-thought answers tailored to the questions, enhancing prediction accuracy.

\begin{table}[!t]
\setlength{\tabcolsep}{8pt}
\centering
\scalebox{0.80}{
\begin{tabular}{lccc}
\toprule
Method                 & LIAR & BBH  & WebNLG \\
\midrule
Zero-shot              & 62.9 & 75.7 & 56.9   \\
Five relevant examples & 65.7 & 78.5 & 57.4   \\
Ours                   & 66.6 & 82.6 & 58.8   \\
\bottomrule
\end{tabular}
}
\caption{Comparison of our method and dynamically selecting five relevant examples using k-nearest neighbors (kNN) clustering in the embedding space.}
\label{tab:cpr_few_shot}
\end{table}

\noindent
\modify{
\textbf{Computational Overhead of Each Component in ERM.}
(1) During the prompt optimization phase, the optimization time mainly comes from the LLM API calls and the memory mechanisms' overhead. 
We measure the average time distribution for each prompt optimization. 
As shown in Table~\ref{tab:computation_overhead_of_memory}, the average time for each optimization LLM API call is 28.161 seconds, while the execution time for the memory mechanisms is 0.062 seconds (we run BGE-M3 on A800 GPU). 
Since memory mechanisms primarily involve the extraction and retrieval of BGE-M3 features, the cost is low, and the execution time overhead is less than 1\%, which can be considered negligible.
(2) During the inference stage, the cost time mainly comes from the LLM API calls and the exemplar retrieval overhead. 
We measure the average time distribution for each prediction stage. 
As shown in Table~\ref{tab:computation_overhead_of_memory}, the average time for LLM API calls is 1.201 seconds, and the average time for exemplar retrieval is 0.029 seconds. 
The time overhead for exemplar retrieval is also negligible. 
}

\begin{table}[!t]
\centering
\setlength{\tabcolsep}{5.4pt}
\scalebox{0.80}{
\begin{tabular}{lcccc}
\toprule
\multirow{2}{*}{Stage} & \multicolumn{2}{c}{\textit{Prompt optimizing}} & \multicolumn{2}{c}{\textit{Inference}} \\
\cmidrule(lr){2-3} \cmidrule(lr){4-5}
& \makecell[c]{\;\;LLM\;\;\\API}  & \makecell[c]{Memory\\Mechanisms} & \makecell[c]{\;\;LLM\;\;\\API} & \makecell[c]{Exemplar\\Retrieval}   \\
\midrule
time~(s)       & 28.161 & 0.062 & 1.201 & 0.029 \\
\bottomrule
\end{tabular}
}
\caption{
\modify{Computational overhead introduced by the memory mechanisms and exemplar retrieval.}
}
\label{tab:computation_overhead_of_memory}
\end{table}

\noindent
\modify{
\textbf{Case Studies.} 
In Tables~\ref{tab:case_study_liar}, \ref{tab:case_study_bbh}, and \ref{tab:case_study_gsm8k}, we present examples of prompt optimization with Feedback Memory on the LIAR, BBH, and GSM8K datasets, respectively. 
Feedback memory effectively combines multiple feedbacks and utilizes them to optimize prompts, preventing the loss of valuable feedbacks during the optimization process. 
This approach results in a stronger optimized prompt.
}

\begin{table*}[!t]
\centering
\scalebox{0.80}{
\begin{tabular}{P{19cm}}
\toprule[1pt]
\textbf{Original Prompt} \\
\hdashline \noalign{\vskip 0.5ex}
Evaluate the accuracy of each statement by determining if it is false (label: Yes) or true (label: No). A statement is false if it contains incorrect information, misleading claims. \\
Evaluation guidelines:\\
1. Misleading or incorrect statistics.\\
2. Incorrect attribution of actions, policies, or quotes.\\
3. Inaccurate reporting of geographical, scientific, or historical facts.\\
4. Overgeneralizations or oversimplifications.\\
\midrule
\textbf{Sampled Feedbacks} \\
\hdashline \noalign{\vskip 0.5ex}
\textsc{Feedback 1} - Assessing Verifiable vs. \textcolor[RGB]{220,0,0}{Unverifiable Statements}: The prompt should aim to clearly distinguish between statements that can be verified with concrete data and those that are more subjective or political in nature. By doing so, it can improve accuracy in classification and ensure a balanced approach to different types of statements. \\
\hdashline \noalign{\vskip 0.5ex}
\textsc{Feedback 2} - \textcolor[RGB]{0,0,220}{Contextual Understanding}: The prompt should enhance its ability to accurately interpret the context and consider the broader background of statements, such as the specifics of Barack Obama's upbringing or global tax comparisons. This will improve its effectiveness in identifying and addressing potential misinformation by ensuring a more comprehensive analysis of the information presented. \\
\hdashline \noalign{\vskip 0.5ex}
\textsc{Feedback 3} - Handling \textcolor[RGB]{0,220,0}{Emotionally Charged Language}: The prompt should be designed to effectively address statements with emotionally charged language or hyperbole. By incorporating explicit instructions for recognizing and managing such language, the model can improve its classification accuracy and better discern the intent and veracity of the information presented. \\
\midrule
\textbf{Optimized Prompt}\\
\hdashline \noalign{\vskip 0.5ex}
Assess the truthfulness of each statement by determining whether it is false (label: Yes) or true (label: No). A statement is deemed false if it contains incorrect information, misleading assertions, or unverifiable claims. Provide a brief explanation for your evaluation, focusing on context, historical records, and specific data points. \\
Evaluation criteria: \\
1. Incorrect or misrepresented statistics.  \\
2. Incorrect attribution of actions, policies, or quotes.   \\ 
3. False reporting of geographical, scientific, or historical facts.   \\ 
4. Ambiguous or \textcolor[RGB]{220,0,0}{unverifiable claims}.  \\ 
5. Overgeneralizations or oversimplifications. \\  
6. Misleading or \textcolor[RGB]{0,220,0}{emotionally charged language}.  \\ 
7. Lack of \textcolor[RGB]{0,0,220}{context affecting the interpretation}. \\
\bottomrule[1pt]
\end{tabular}
}
\caption{
\modify{Intermediate prompt optimized by Feedback Memory on the LIAR dataset.}
}
\label{tab:case_study_liar}
\end{table*}
\begin{table*}[!t]
\centering
\scalebox{0.80}{
\begin{tabular}{P{19cm}}
\toprule[1pt]
\textbf{Original Prompt} \\
\hdashline \noalign{\vskip 0.5ex}
You are given a set of instructions to follow. Determine if you will return to the starting point after executing all the instructions. Consider the spatial relationships and movements, including turns and steps in various directions. The possible answers are:\\
- Yes\\
- No\\
\midrule
\textbf{Sampled Feedbacks} \\
\hdashline \noalign{\vskip 0.5ex}
\textsc{Feedback 1} - Add Details: Clearly specify the types of movements and turns, \textcolor[RGB]{220,0,0}{like left, right, forward, and backward}. \\
\hdashline \noalign{\vskip 0.5ex}
\textsc{Feedback 2} - Direction Guidance: Emphasize handling direction changes and ensure the \textcolor[RGB]{0,220,0}{"always face forward"} concept is clear. \\
\hdashline \noalign{\vskip 0.5ex}
\textsc{Feedback 3} - \textcolor[RGB]{0,0,220}{Cumulative Effect}: Instruct users to calculate the cumulative effect of all movements and turns to determine the final position. \\
\midrule
\textbf{Optimized Prompt}\\
\hdashline \noalign{\vskip 0.5ex}
You are provided with a sequence of movement instructions. Your objective is to determine if following these instructions will bring you back to the starting point. Consider every movement and turn mentioned, including steps to the \textcolor[RGB]{220,0,0}{left, right, forward, and backward}. The directive \textcolor[RGB]{0,220,0}{"Always face forward"} implies maintaining your original direction unless specified to turn. \textcolor[RGB]{0,0,220}{Accumulate the total effect} of all movements and turns to determine the final position. The possible results are: \\
- Yes \\
- No \\
\bottomrule[1pt]
\end{tabular}
}
\caption{
\modify{Intermediate prompt optimized by Feedback Memory on the BBH dataset.}
}
\label{tab:case_study_bbh}
\end{table*}
\begin{table*}[!t]
\centering
\scalebox{0.80}{
\begin{tabular}{P{19cm}}
\toprule[1pt]
\textbf{Original Prompt} \\
\hdashline \noalign{\vskip 0.5ex}
Approach this problem methodically by following these steps: Ensure that you double-check your calculations and verify your results for accuracy. Consider each part of the problem carefully, breaking it down into manageable steps, and pay attention to any special conditions, sequential events, or additional constraints that could impact the accuracy of your solution. Focus on numerical variations and ensure all given numbers are correctly interpreted and used in the calculations.\\
\midrule
\textbf{Sampled Feedbacks} \\
\hdashline \noalign{\vskip 0.5ex}
\textsc{Feedback 1} - \textcolor[RGB]{220,0,0}{Structured Approach}: The original prompt would benefit from a more structured approach, outlining specific steps like interpretation, breaking down the problem, calculation, and verification to enhance clarity and guidance. \\
\hdashline \noalign{\vskip 0.5ex}
\textsc{Feedback 2} - \textcolor[RGB]{0,220,0}{Emphasize Relationships}: Highlight the importance of understanding relationships, conditions, and constraints within the problem to ensure accurate interpretation and solution. \\
\hdashline \noalign{\vskip 0.5ex}
\textsc{Feedback 3} - \textcolor[RGB]{0,0,220}{Verification Focus}: Stress the need for a thorough verification process, ensuring each step logically follows and the final result is contextually sound, encouraging recalculation if needed. \\
\midrule
\textbf{Optimized Prompt}\\
\hdashline \noalign{\vskip 0.5ex}
Approach this problem methodically by following these steps: \\
\textcolor[RGB]{220,0,0}{1.} **\textcolor[RGB]{0,220,0}{Interpretation}:** Carefully read and interpret the problem statement. Pay close attention to the relationships, conditions, constraints, and sequence of events described. Identify key quantities and their interrelationships.\\
\textcolor[RGB]{220,0,0}{2.} **Break Down:** Break the problem into manageable steps. Identify the calculations required for each step and the sequence in which to perform them. Ensure you understand how the different parts of the problem connect.\\
\textcolor[RGB]{220,0,0}{3.} **Calculation:** Perform the calculations step-by-step. Ensure that each calculation is based on the correct interpretation of the problem's conditions. Be meticulous with numerical values, units, and any given constraints.\\
\textcolor[RGB]{220,0,0}{4.} **\textcolor[RGB]{0,0,220}{Verification}:** Double-check your calculations. Verify that each step logically follows from the previous one and that the final result makes sense in the context of the problem. Recalculate if necessary to ensure accuracy. \\
\bottomrule[1pt]
\end{tabular}
}
\caption{
\modify{Intermediate prompt optimized by Feedback Memory on the GSM8K dataset.}
}
\label{tab:case_study_gsm8k}
\end{table*}

\section{More Discussion}

\noindent
\modify{
\textbf{Potential Benefit of Memory Mechanism.}
In practical applications, prompt optimization might not always generate a prompt that meets human expectations. 
Thanks to our proposed memory mechanism, we can easily incorporate human intervention. 
Specifically, we have explored the following two aspects: \\
1) When the model struggles with a particular exemplar and fails to resolve it successfully, we can trigger human intervention to verify whether the answer for that exemplar is correct, thereby correcting noisy labels or considering the addition of a chain of thought for that exemplar. \\
2) When the prompt fails to achieve the desired effect, we can introduce human feedbacks to improve prompt optimization.
}

\noindent
\modify{
\textbf{Analysis about Performance Improvements.}
The performance improvement of our method stems from both indirect feedbacks and direct exemplars. \\
1) Exemplar-guided Reflection enables the \promptoptimizer{} to generate strong feedbacks. 
The Feedback Memory collects potentially valuable feedbacks, avoiding the issue of feedback forgetting that may occur with sequential optimization of multiple feedbacks. By considering multiple feedbacks simultaneously, it enhances the language model's ability to retain information, leading to the generation of better prompts and therefore indirectly boosting performance. \\
2) Our proposed Exemplar Factory identifies exemplars that are more targeted to specific questions, therefore directly enhancing the prediction performance.
}

\section{Meta-Prompt}

Here are the meta-prompts we used in Section~\ref{sec:method}.

\begin{figure*}[h]
\footnotesize
    \centering
        \begin{tcolorbox}[colback=gray!2!white, colframe=gray!10!black]
        \texttt{I'm trying to write and complete a zero-shot classifier prompt from difficult or erroneous examples, `text' field means model input, `label' field means true label.\\
My current prompt is: \\
\{curr\_prompt\} \\
But this prompt gets the following examples wrong: \\
\{error\_samples\} \\
To improve my understanding and performance, I would like to identify \{num\_exemplar\} typical examples from the above cases where the current prompt fails. \\
These examples should be diverse to cover a range of different issues. \\
For each example, provide the following format in JSON and wrap each example with <key\_example> and </key\_example>: \\
<key\_example> \\
\{ \\
  ``text": ``\{\{text\}\}",\\
  ``label": ``\{\{label\}\}",\\
  ``solution": ``How to solve this problem step-by-step to get a more accurate answer."\\
\}\\
</key\_example>\\
After identifying these \{num\_exemplar\} typical examples, please provide \{num\_feedbacks\} reasons why the prompt could have gotten these examples wrong. Wrap each reason with <feedback> and </feedback>.\\
        }
    \end{tcolorbox}
      \caption{Intructive reflection meta-prompt.}
      \label{fig:appendix_meta_prompt_reflection}
\end{figure*}
\begin{figure*}[h]
\footnotesize
    \centering
        \begin{tcolorbox}[colback=gray!2!white, colframe=gray!10!black]
        \texttt{I'm trying to write and complete a zero-shot classifier prompt from difficult or erroneous examples, `text' field means model input, `label' field means true label.\\
My current prompt is:\\
\{prompt\} \\
But this prompt gets the following examples wrong:\\
\{error\_samples\}\\
Based on these examples the problem with this prompt is that: \\
\{feedback\} \\
Based on the above information, I refine the prompt to make the model predict correctly.  \\
The refined prompt is wrapped with <prompt> and </prompt>, less that 512 words:
        }
    \end{tcolorbox}
      \caption{Optimization meta-prompt.}
      \label{fig:appendix_meta_prompt_optimization}
\end{figure*}
\begin{figure*}[h]
\footnotesize
    \centering
        \begin{tcolorbox}[colback=gray!2!white, colframe=gray!10!black]
        \texttt{I'm trying to write and complete a zero-shot classifier prompt from difficult or erroneous examples, `text' field means model input, `label' field means true label.\\
My current prompt is:\\
\{prompt\} \\
Here are some examples of issues and their labels:\\
\{error\_samples\}\\
Here are some suggestions for improving the prompt: \\
\{feedbacks\} \\
Based on the above information, I refine the prompt to make the model predict correctly.   \\
The refined prompt is wrapped with <prompt> and </prompt>, less that 512 words:
        }
    \end{tcolorbox}
      \caption{Retrieval optimization meta-prompt.}
      \label{fig:appendix_meta_prompt_retrieval_optimization}
\end{figure*}

\section{Additional Result}
\label{sec:addition_result}

Here, we present the initial prompt, the ProTeGi-optimized prompt, and \shortname{}-optimized prompt across different tasks.

\begin{figure*}[h]
\footnotesize
    \centering
        \begin{tcolorbox}[colback=gray!2!white, colframe=gray!10!black]
        \textbf{Initial Prompt (Empty)}\\
        \texttt{
\#\# Task \\
Determine whether the Statement is a lie (Yes) or not (No) based on the Context and other information. \\
\#\# Output format \\
Answer Yes or No as labels. \\
\#\# Prediction \\
Text: \{input\} \\
Label: \\
        }
    \end{tcolorbox}
      \caption{Initial prompt of the LIAR dataset.}
      \label{fig:appendix_liar_instruction_init}
\end{figure*}

\begin{figure*}[h]
\footnotesize
    \centering
        \begin{tcolorbox}[colback=gray!2!white, colframe=gray!10!black]
        \textbf{ProTeGi Optimized Prompt}\\
        \texttt{
\#\# Task \\
Evaluate the Statement below using the provided Context and ascertain its factual accuracy (Yes, it is false or misleading) or accuracy (No, it is not false or misleading). Follow these steps for your evaluation: \\
1. Confirm the factual accuracy of the Statement by referencing the given Context and relevant background information. \\
2. Take into account the job title, state, and political affiliation of the speaker to gauge their perspective and potential bias. \\
3. Assess the plausibility and logical coherence of the Statement. \\
4. Verify the Statement against established facts and data as necessary. \\
5. Evaluate whether the Statement, even if factually accurate, is presented in a misleading or hyperbolic manner. \\
Statement: A study of private bail bond systems showed that Wisconsin has a higher no-show rate than other states of defendants skipping court appearances. \\
Job title: Wisconsin Assembly speaker \\
State: Wisconsin \\
Party: Republican \\
Context: an interview \\
\#\# Output format \\
Answer Yes or No as labels. \\
\#\# Prediction \\
Text: \{input\} \\
Label: \\
     } 
    \end{tcolorbox}
      \caption{ProTeGi optimized prompt of the LIAR dataset.}
      \label{fig:appendix_liar_instruction_protegi}
\end{figure*}

\begin{figure*}[h]
\footnotesize
    \centering
        \begin{tcolorbox}[colback=gray!2!white, colframe=gray!10!black]
     \textbf{\shortname{} Optimized Prompt}\\
     \texttt{
\#\# Task \\
You are tasked with determining the factual accuracy of statements based on their content, context, and widely accepted facts. Your goal is to decide whether the statement is false (``Yes") or true (``No"). For each example, you will be provided with: \\
1. **Statement**: The statement to be evaluated. \\
2. **Job title**: The job title of the person who made the statement (if available). \\
3. **State**: The state associated with the person who made the statement (if available). \\
4. **Party**: The political party of the person who made the statement (if available). \\
5. **Context**: The situation in which the statement was made, including any relevant background information. \\
Instructions: \\
1. **Evaluate the statement** based on verifiability and supporting evidence from **multiple reliable sources**. \\
2. **Cross-reference the statement** with verifiable data and widely accepted facts. \\ 
3. **Consider the context** in which the statement was made, including legislative, historical, and situational nuances. \\ 
4. **Ignore the political affiliation** and focus solely on the factual accuracy of the statement. \\
5. **If a statement is vague, lacks concrete details, or cannot be verified with reliable sources, answer ``Yes."** \\
6. **If a statement is partially true but omits crucial context or presents facts misleadingly, answer ``Yes."** \\
7. **If a statement is true and well-supported by reliable evidence, answer ``No."**\\
8. **Pay special attention** to statements with mixed truths; if any part of the statement is misleading, answer ``Yes." \\ 
9. **If a statement is statistically accurate but requires nuanced interpretation or context to be fully understood, answer ``No."** \\
10. **Be mindful of hyperbolic, rhetorical, or satirical elements**. If the core factual content is accurate and verifiable, answer ``No." If hyperbole or rhetoric leads to a misleading impression, answer ``Yes." \\
11. **Prioritize factual accuracy** and ensure your decision is based on concrete evidence and context. \\
12. **For statements with mixed or nuanced truths**, focus on whether the core message is accurate. If the core message is misleading or omits critical context, answer ``Yes." If the core message is accurate despite requiring nuanced interpretation, answer ``No." \\
Example 1: \\
- Statement: ``Every 28 hours an unarmed black person is shot by a cop." \\
- Job title: Activist \\
- State: California \\
- Party: none \\
- Context: a speech at a rally \\
- Answer: Yes \\ 
Example 2: \\
- Statement: ``Congressman Renacci is under FBI investigation." \\
- Job title: Politician \\ 
- State: Ohio \\
- Party: republican \\
- Context: a news interview \\
- Answer: Yes \\
Example 3: \\
- Statement: ``You can't build a Christian church in Saudi Arabia."\\
- Job title: Radio/TV host \\ 
- State: \\
- Party: none \\
- Context: a broadcast on the Sean Hannity radio show \\
- Answer: No \\
\#\# Output format \\
Answer Yes or No as labels. \\
\#\# Prediction \\
Text: \{input\} \\
Label:}
    \end{tcolorbox}
      \caption{\shortname{} optimized prompt of the LIAR dataset.}
      \label{fig:appendix_liar_instruction_ours}
\end{figure*}

\begin{figure*}[h]
\footnotesize
    \centering
        \begin{tcolorbox}[colback=gray!2!white, colframe=gray!10!black]
        \textbf{Initial Prompt (Empty)}\\
        \texttt{
\#\# Task \\
If you follow these instructions, do you return to the starting point? \\
\#\# Output format \\
Answer Yes or No as labels. \\
\#\# Prediction \\
Text: \{input\} \\
Label: \\
        }
    \end{tcolorbox}
      \caption{Initial prompt of the BBH dataset.}
      \label{fig:appendix_bbh_navigate_instruction_init}
\end{figure*}

\begin{figure*}[h]
\footnotesize
    \centering
        \begin{tcolorbox}[colback=gray!2!white, colframe=gray!10!black]
        \textbf{ProTeGi Optimized Prompt}\\
        \texttt{
\#\# Task \\
If you follow these instructions, do you return to the starting point? \\
\#\# Output format \\
Answer Yes or No as labels. \\
\#\# Prediction \\
Text: \{input\} \\
Label: \\
     } 
    \end{tcolorbox}
      \caption{ProTeGi optimized prompt of the BBH dataset.}
      \label{fig:appendix_bbh_navigate_instruction_protegi}
\end{figure*}

\begin{figure*}[h]
\footnotesize
    \centering
        \begin{tcolorbox}[colback=gray!2!white, colframe=gray!10!black]
     \textbf{\shortname{} Optimized Prompt}\\
     \texttt{
\#\# Task \\
You are provided with a sequence of movement instructions. Your objective is to determine if following these instructions will bring you back to the starting point. Consider every movement and turn mentioned, including steps to the left, right, forward, and backward. The directive ``Always face forward" implies maintaining your original direction unless specified to turn. Accumulate the total effect of all movements and turns to determine the final position. The possible results are: \\
- Yes \\
- No \\
Consider these examples: \\
1. Instructions: Always face forward. Move 7 steps forward. Move 7 steps backward. Move 10 steps backward. Move 10 steps forward. \\
   Answer: Yes \\
2. Instructions: Move 8 steps. Move 1 step. Turn left. Turn left. Move 9 steps. \\
   Answer: Yes \\
3. Instructions: Move 4 steps. Move 2 steps. Turn around. Move 6 steps. \\
   Answer: Yes \\
4. Instructions: Move 3 steps. Turn left. Turn left. Turn around. Move 5 steps. Turn around. Move 5 steps. \\
   Answer: No \\
5. Instructions: Move 6 steps. Move 10 steps. Move 10 steps. Move 1 step. Move 7 steps. Move 5 steps. Turn around. \\
   Answer: No \\
Now, review the following instructions and determine if you will return to the starting point:\\
\#\# Output format \\
Answer Yes or No as labels. \\
\#\# Prediction \\
Text: \{input\} \\
Label:}
    \end{tcolorbox}
      \caption{\shortname{} optimized prompt of the BBH dataset.}
      \label{fig:appendix_bbh_navigate_instruction_ours}
\end{figure*}

\begin{figure*}[h]
\footnotesize
    \centering
        \begin{tcolorbox}[colback=gray!2!white, colframe=gray!10!black]
        \textbf{Initial Prompt (Empty)}\\
        \texttt{
\#\# Task \\
Is the following text hate speech? \\
\#\# Output format \\
Answer Yes or No as labels. \\
\#\# Prediction \\
Text: \{input\} \\
Label: \\
        }
    \end{tcolorbox}
      \caption{Initial prompt of the ETHOS dataset.}
      \label{fig:appendix_ethos_instruction_init}
\end{figure*}

\begin{figure*}[h]
\footnotesize
    \centering
        \begin{tcolorbox}[colback=gray!2!white, colframe=gray!10!black]
        \textbf{ProTeGi Optimized Prompt}\\
        \texttt{
\#\# Task \\
Input: Please determine whether the provided text constitutes hate speech. Hate speech encompasses any expression that belittles, discriminates against, or dehumanizes individuals or groups based on attributes such as race, religion, ethnic background, sexual orientation, disability, or gender. Analyze if the text contains harmful stereotypes, derogatory language, or promotes violence or discrimination, including subtle, coded, sarcastic, or ironic language. Take into account the context and any potentially harmful intentions behind the words, even if they are not overtly stated. Pay particular attention to implicit forms of hate speech that may not be explicitly offensive but still perpetuate harmful stereotypes or intentions. \\
Examples: \\
- ``Men race is a subordinate of women": Yes, this is hate speech. \\
- ``I love deaf people. I can scream on them and I can scold them": Yes, this is hate speech. \\
- ``Don't hit Women. Just fuck them and make them pregnant.": Yes, this is hate speech. \\
- ``Autists is why presidents like trump are getting elected": Yes, this is hate speech. \\
Now, please evaluate the following text: \\
\#\# Output format \\
Answer Yes or No as labels. \\
\#\# Prediction \\
Text: \{input\} \\
Label: \\
     } 
    \end{tcolorbox}
      \caption{ProTeGi optimized prompt of the ETHOS dataset.}
      \label{fig:appendix_ethos_instruction_protegi}
\end{figure*}

\begin{figure*}[h]
\footnotesize
    \centering
        \begin{tcolorbox}[colback=gray!2!white, colframe=gray!10!black]
     \textbf{\shortname{} Optimized Prompt}\\
     \texttt{
\#\# Task \\
You are an expert in identifying hate speech. Hate speech includes any expression, gesture, conduct, writing, or display that incites violence or prejudicial action against a particular individual or group or disparages or intimidates a protected individual or group. This includes derogatory comments about race, religion, ethnicity, sexual orientation, disability, gender, nationality, medical conditions (such as autism), and more. Hate speech can target individuals or entire communities and can include threats, slurs, dehumanizing language, and claims of superiority or inferiority of one gender or group over another. \\
Given the following text, determine whether it qualifies as hate speech or not. Answer ``Yes" if it is hate speech and ``No" if it is not. \\
Consider the context, implicit meanings, and any underlying biases or derogatory generalizations in the text. Pay attention to any language that may incite violence, express hate, promote discrimination, or dehumanize specific groups, including nationalities, religions, disabilities, medical conditions, or other protected characteristics. Be sensitive to sarcasm, coded language, indirect expressions of hate, and claims of gender superiority or inferiority. Hate speech can be directed at any group, including white people and women. Remember to include exclusionary practices, ableist language, and derogatory comments about medical conditions as potential hate speech. \\
*** Note: ** Scientific debates or discussions that do not contain derogatory language or intentions to harm specific groups should not be considered hate speech. Focus on identifying language that explicitly or implicitly promotes hate, violence, or dehumanization. \\
Examples of hate speech include but are not limited to: \\
- Racist slurs or statements promoting racial superiority. \\
- Comments that dehumanize or demean individuals based on their religion. \\
- Statements advocating violence or harm against a specific group. \\
- Derogatory terms or phrases targeting sexual orientation. \\
- Ableist language such as ``retard," ``mentally ill," or any other term that demeans disabilities. \\
Examples: \\
Text: ``Transgenderism is a DISEASE. Transgenders are an abomination. Not only that, but they're also lying to themselves. You CANNOT change your DNA from male to female, or from female to male. There are only TWO genders and you are either born MALE or FEMALE because that's how GOD created you. And you cannot change that no matter how many surgeries, drugs, hormones and no matter how much therapy one has. FACT." \\
Label: ``Yes" \\
Text: ``Transgenders have to protect each other, not fight each other." \\
Label: ``No" \\
Text: ``Gayness should be removed from the face of the earth." \\
Label: ``Yes" \\
... \\
\#\# Output format \\
Answer Yes or No as labels. \\
\#\# Prediction \\
Text: \{input\} \\
Label:}
    \end{tcolorbox}
      \caption{\shortname{} optimized prompt of the ETHOS dataset.}
      \label{fig:appendix_ethos_instruction_ours}
\end{figure*}

\begin{figure*}[h]
\footnotesize
    \centering
        \begin{tcolorbox}[colback=gray!2!white, colframe=gray!10!black]
        \textbf{Initial Prompt (Empty)}\\
        \texttt{
\#\# Task \\
Is this tweet sarcastic? \\
\#\# Output format \\
Answer Yes or No as labels. \\
\#\# Prediction \\
Text: \{input\} \\
Label: \\
        }
    \end{tcolorbox}
      \caption{Initial prompt of the ArSarcasm dataset.}
      \label{fig:appendix_arsarcasm_instruction_init}
\end{figure*}

\begin{figure*}[h]
\footnotesize
    \centering
        \begin{tcolorbox}[colback=gray!2!white, colframe=gray!10!black]
        \textbf{ProTeGi Optimized Prompt}\\
        \texttt{
\#\# Task \\
Kindly assess the provided tweet to determine if it uses sarcasm. Consider the cultural nuances, linguistic cues, and overall tone in your evaluation. Offer a comprehensive explanation of your findings: \\
Text: ``<arabic sentences not supported for display>" \\
Conclusion: Yes. The phrase ``<arabic sentences not supported for display>" (which translates to ``a sheep against the enemy, a lion against the elderly and children") employs sarcasm to critique someone for showing courage only towards those who are vulnerable. \\
\#\# Output format \\
Answer Yes or No as labels. \\
\#\# Prediction \\
Text: \{input\} \\
Label: \\
     } 
    \end{tcolorbox}
      \caption{ProTeGi optimized prompt of the ArSarcasm dataset.}
      \label{fig:appendix_arsarcasm_instruction_protegi}
\end{figure*}

\begin{figure*}[h]
\footnotesize
    \centering
        \begin{tcolorbox}[colback=gray!2!white, colframe=gray!10!black]
     \textbf{\shortname{} Optimized Prompt}\\
     \texttt{
\#\# Task \\
Analyze the following tweet to determine if it is sarcastic. Sarcasm often involves saying the opposite of what one means and may contain elements of irony, exaggeration, mockery, or complex emotional undertones. Carefully consider the context, including cultural, political, and social references, which can carry implicit sarcastic undertones in Arabic tweets. Examine the tweet for subtle clues such as understatement, dry humor, and nuanced emotional tone that could indicate sarcasm. \\
Key points to consider: \\
- **Exaggeration:** Look for statements that sound overly dramatic or extreme. \\
- **Irony:** Identify instances where the intended meaning is the opposite of the literal wording. \\
- **Contradictory Statements:** Detect inconsistencies within the tweet itself. \\
- **Cultural, Political, and Social Nuances:** Recognize idioms, cultural references, and politically or socially charged statements that suggest sarcasm. \\
- **Emotional Tone:** Pay attention to signals like bitterness, frustration, mockery, or exaggerated enthusiasm, which are key indicators of sarcasm. \\
- **Subtle Clues:** Look for understated comments, dry humor, or nuanced emotional expressions that may indicate sarcasm. This includes seemingly positive statements with a negative context or vice versa, and overly enthusiastic remarks that may carry an underlying negative sentiment. \\
Examples to guide your analysis: \\
1. ``<arabic sentences not supported for display>" – Yes \\
2. ``<arabic sentences not supported for display>" – No \\
3. ``<arabic sentences not supported for display>" – Yes \\
Now, decide if the given tweet is sarcastic and answer with either ``Yes" or ``No". \\
\#\# Output format \\
Answer Yes or No as labels. \\
\#\# Prediction \\
Text: \{input\} \\
Label:}
    \end{tcolorbox}
      \caption{\shortname{} optimized prompt of the ArSarcasm dataset.}
      \label{fig:appendix_arsarcasm_instruction_ours}
\end{figure*}

\begin{figure*}[h]
\footnotesize
    \centering
        \begin{tcolorbox}[colback=gray!2!white, colframe=gray!10!black]
        \textbf{Initial Prompt (Empty)}\\
        \texttt{
\#\# Task \\
Write the following triples as fluent English text. \\
\#\# Prediction \\
\{input\} \\
Answer:
        }
    \end{tcolorbox}
      \caption{Initial prompt of the WebNLG dataset.}
      \label{fig:appendix_webnlg_instruction_init}
\end{figure*}

\begin{figure*}[h]
\footnotesize
    \centering
        \begin{tcolorbox}[colback=gray!2!white, colframe=gray!10!black]
        \textbf{ProTeGi Optimized Prompt}\\
        \texttt{
\#\# Task\\
You are given a set of triples that need to be converted into coherent and fluent English sentences. Each triple consists of a subject, predicate, and object. Your task is to accurately convey the information from these triples into well-formed sentences. Ensure that the sentences are complete, grammatically correct, and clearly express the relationships provided in the triples.  \\
Guidelines: \\
1. Combine related triples into a single sentence where appropriate. \\
2. Use synonyms and variations to avoid repetition, but ensure the meaning remains clear and accurate. \\ 
3. Incorporate all relevant information for each subject within the same sentence or group of sentences. \\
4. Maintain the context and coherence of the information while ensuring the sentences flow naturally. \\
5. Be mindful of the sequence of information to enhance readability and understanding. \\
6. Clearly differentiate between simple and more complex relationships to fully capture the depth of the information provided. Pay particular attention to hierarchical relationships or ownership, clearly distinguishing between entities such as manufacturers, subsidiaries, and divisions. \\
\#\# Prediction \\
\{input\} \\
Answer:
     }
    \end{tcolorbox}
      \caption{ProTeGi optimized prompt of the WebNLG dataset.}
      \label{fig:appendix_webnlg_instruction_protegi}
\end{figure*}

\begin{figure*}[h]
\footnotesize
    \centering
        \begin{tcolorbox}[colback=gray!2!white, colframe=gray!10!black]
     \textbf{\shortname{} Optimized Prompt}\\
     \texttt{
\#\# Task \\
Convert the following triples into coherent and fluent English sentences. Ensure that all relationships and attributes are accurately conveyed. When multiple associations or attributes are involved, break down the information into smaller, logical sentences to maintain clarity. \\ 
Example 1: \\
Triples: \\
Anders\_Osborne | associatedBand/associatedMusicalArtist | Billy\_Iuso \\
Anders\_Osborne | associatedBand/associatedMusicalArtist | Tab\_Benoit \\
Anders\_Osborne | genre | Rock\_music \\
Anders\_Osborne | associatedBand/associatedMusicalArtist | Galactic \\
Output: \\
Rock musician Anders Osborne has worked with the band Galactic and also with the musical artists Tab Benoit and Billy Iuso. \\
Example 2:\\
Triples:\\
Twilight\_(band) | genre | Black\_metal\\
Aaron\_Turner | associatedBand/associatedMusicalArtist | Twilight\_(band)\\
Aaron\_Turner | associatedBand/associatedMusicalArtist | House\_of\_Low\_Culture\\
Aaron\_Turner | instrument | Electric\_guitar\\
Black\_metal | musicFusionGenre | Death\_metal\\
Output:\\
Aaron Turner plays the electric guitar and performed with Twilight, a black metal band, and House of Low Culture. Black metal is an element of the fusion genre death metal.\\
Example 3:\\
Triples:\\
Baked\_Alaska | mainIngredient | ``Meringue, ice cream, sponge cake or Christmas pudding"\\
Baked\_Alaska | country | ``France, United States or China"\\
Baked\_Alaska | region | ``Paris, New York or Hong Kong"\\
Baked\_Alaska | ingredient | Meringue\\
...\\
Output:\\
Baked Alaska has the main ingredients of meringue, ice cream, and sponge cake (or Christmas pudding). It is found in France, the US, China, Hong Kong, New York, and Paris.\\
\#\# Prediction \\
\{input\} \\
Answer:
     }
    \end{tcolorbox}
      \caption{\shortname{} optimized prompt of the WebNLG dataset.}
      \label{fig:appendix_webnlg_instruction_ours}
\end{figure*}
\begin{figure*}[h]
\footnotesize
    \centering
        \begin{tcolorbox}[colback=gray!2!white, colframe=gray!10!black]
        \textbf{Initial Prompt (Empty)}\\
        \texttt{
\#\# Task \\
Solve the math problem. \\
\#\# Prediction \\
Text: \{input\} \\
Label: \\
        }
    \end{tcolorbox}
      \caption{Initial prompt of the GSM8K dataset.}
      \label{fig:appendix_gsm8k_instruction_init}
\end{figure*}

\begin{figure*}[h]
\footnotesize
    \centering
        \begin{tcolorbox}[colback=gray!2!white, colframe=gray!10!black]
        \textbf{ProTeGi Optimized Prompt}\\
        \texttt{
\#\# Task \\
Read the following problem carefully and perform the necessary mathematical calculations to find the correct numerical answer. \\
\#\# Prediction \\
Text: \{input\} \\
Label: \\
     } 
    \end{tcolorbox}
      \caption{ProTeGi optimized prompt of the GSM8K dataset.}
      \label{fig:appendix_gsm8k_instruction_protegi}
\end{figure*}

\begin{figure*}[h]
\footnotesize
    \centering
        \begin{tcolorbox}[colback=gray!2!white, colframe=gray!10!black]
     \textbf{\shortname{} Optimized Prompt}\\
     \texttt{
\#\# Task \\
Approach this problem methodically by following these steps:  \\
1. **Interpretation:** Carefully read and interpret the problem statement. Pay close attention to the relationships, conditions, constraints, and sequence of events described. Identify key quantities and their interrelationships. \\
2. **Break Down:** Break the problem into manageable steps. Identify the calculations required for each step and the sequence in which to perform them. Ensure you understand how the different parts of the problem connect. \\
3. **Calculation:** Perform the calculations step-by-step. Ensure that each calculation is based on the correct interpretation of the problem's conditions. Be meticulous with numerical values, units, and any given constraints. \\
4. **Verification:** Double-check your calculations. Verify that each step logically follows from the previous one and that the final result makes sense in the context of the problem. Recalculate if necessary to ensure accuracy. \\
Refer to the following examples for guidance: \\
Example 1: \\
Text: ``At the burger hut, you can buy a burger for \$5, french fries for \$3, and a soft drink for \$3. If you order a special burger meal, you get all 3 of these food items for \$9.50. A kid’s burger is \$3, a kid’s french fries are \$2, and a kid's juice box is \$2. They also have a kids meal of all 3 kids' food items for \$5. Mr. Parker buys 2 burger meals for his wife and himself. He also buys 2 burger meals and 2 kid's meals for his 4 children. How much money does Mr. Parker save by buying the 6 meals versus buying the individual food items?" \\
Solution Steps: \\
- Calculate the individual cost of each adult meal: \$5 + \$3 + \$3 = \$11. \\
- Total cost for 4 adult meals: 4 * \$11 = \$44. \\
- Calculate the cost of each kid’s meal: \$3 + \$2 + \$2 = \$7. \\
- Total cost for 2 kids’ meals: 2 * \$7 = \$14.\\ 
- Total cost without meal deals: \$44 + \$14 = \$58. \\
- Cost with meal deals: 4 * \$9.50 (adult meals) + 2 * \$5 (kids’ meals) = \$38 + \$10 = \$48. \\
- Total savings: \$58 - \$48 = \$10. \\
Label: 10.\\
Example 2: \\
Text: ``Liam wants to go to Paris, but first, he has to pay his bills. His trip costs \$7,000, and his bills cost \$3,500. Knowing that Liam has saved \$500/month for 2 years, how much money will he have left after paying his bills?"
Solution Steps: \\
- Total savings: \$500 * 24 months = \$12,000. \\
- Total expenses (trip + bills): \$7,000 + \$3,500 = \$10,500. \\
- Money left after expenses: \$12,000 - \$10,500 = \$1,500.\\
Label: 1500.\\
Example 3:\\
Text: ``Steve has a bank account that earns 10\% interest every year. He puts \$100 in it, and then \$10 each year. How much money is in it after two years?" \\
Solution Steps: \\
- First year: \$100 * 1.10 + \$10 = \$120. \\
- Second year: \$120 * 1.10 + \$10 = \$142. \\
Label: 142. \\
Use these examples as a guide to solve your problem. Carefully verify each step, consider any numerical variations, and ensure all calculations align with the problem's conditions. Once you have your solution, review it to confirm its validity in the context of the problem.\\
\#\# Prediction \\
Text: \{input\} \\
Label:}
    \end{tcolorbox}
      \caption{\shortname{} optimized prompt of the GSM8K dataset.}
      \label{fig:appendix_gsm8k_instruction_ours}
\end{figure*}

\begin{figure*}[h]
\footnotesize
    \centering
        \begin{tcolorbox}[colback=gray!2!white, colframe=gray!10!black]
        \textbf{Initial Prompt (Empty)}\\
        \texttt{
\#\# Task \\
Solve the problem. \\
\#\# Prediction \\
Text: \{input\} \\
Label: \\
        }
    \end{tcolorbox}
      \caption{Initial prompt of the WSC dataset.}
      \label{fig:appendix_wsc_instruction_init}
\end{figure*}

\begin{figure*}[h]
\footnotesize
    \centering
        \begin{tcolorbox}[colback=gray!2!white, colframe=gray!10!black]
        \textbf{ProTeGi Optimized Prompt}\\
        \texttt{
\#\# Task \\
Carefully read the provided text and identify the entity that the pronoun in the text refers to. Take into account the context, including relationships and actions described. Select the correct option (A or B) that corresponds to the referent of the pronoun. \\
For instance: \\
- Examine actions that might indicate which entity is being referred to. \\
- Consider the logical flow of events. \\
- Notice descriptions and the relative positioning of the entities. \\
Text: ``The sack of potatoes had been placed below the bag of flour, so it had to be moved first. What does the pronoun ``it" refer to? \\
(A) The sack of potatoes \\
(B) The bag of flour" \\
Answer: (B) \\
Text: ``George got free tickets to the play, but he gave them to Eric, because he was particularly eager to see it. What does the pronoun ``he" refer to? \\
(A) George \\
(B) Eric" \\
Answer: (B) \\
Text: ``It was a summer afternoon, and the dog was sitting in the middle of the lawn. After a while, it got up and moved to a spot under the tree, because it was cooler. What does the pronoun ``it" refer to? \\
(A) The dog \\
(B) The spot under the tree" \\
Answer: (B) \\
Text: ``The sculpture rolled off the shelf because it wasn't level. What does the pronoun ``it" refer to ? \\
(A) The sculpture  \\
(B) The shelf" \\
Answer: (B) \\
\#\# Prediction \\
Text: \{input\} \\
Label: \\
     } 
    \end{tcolorbox}
      \caption{ProTeGi optimized prompt of the WSC dataset.}
      \label{fig:appendix_wsc_instruction_protegi}
\end{figure*}

\begin{figure*}[h]
\footnotesize
    \centering
        \begin{tcolorbox}[colback=gray!2!white, colframe=gray!10!black]
     \textbf{\shortname{} Optimized Prompt}\\
     \texttt{
\#\# Task \\
You will be given a sentence or a pair of sentences containing one or more pronouns. Your task is to identify the noun or noun phrase that each pronoun most logically refers to, based on context, causality, descriptive details, and common sense reasoning. Carefully analyze the sentences and use your understanding of typical human behavior, relationships, and world knowledge to determine the correct antecedent for each pronoun. \\
Consider the following guiding principles: \\
1. **Influence and Causality**: Who or what is causing an action or effect? \\
2. **Descriptive Context**: What descriptive details precede or follow the pronoun? \\
3. **Actions and Reactions**: Who is performing or receiving an action? \\
4. **Contextual Dependencies**: Use background knowledge and the usual roles in interactions to resolve pronouns accurately. \\
Examples: \\
1. ``Steve follows Fred's example in everything. He influences him hugely. What does the pronoun `He' refer to?" \\
(A) Steve \\
(B) Fred \\
Answer: (B) Fred \\
2. ``Pete envies Martin because he is very successful. What does the pronoun `he' refer to?" \\
(A) Pete \\
(B) Martin \\
Answer: (B) Martin \\
3. ``Sid explained his theory to Mark but he couldn't convince him. What does the pronoun `he' refer to?" \\
(A) Sid \\
(B) Mark \\
Answer: (A) Sid \\
4. ``The fish ate the worm. It was tasty. What does the pronoun `It' refer to?" \\
(A) The fish \\
(B) The worm \\
Answer: (B) The worm \\
Analyze each sentence and select the option that best fits the context and your general knowledge. \\
\#\# Prediction \\
Text: \{input\} \\
Label:}
    \end{tcolorbox}
      \caption{\shortname{} optimized prompt of the WSC dataset.}
      \label{fig:appendix_wsc_instruction_ours}
\end{figure*}

\end{document}